\documentclass{article}

\PassOptionsToPackage{numbers,sort&compress}{natbib}

\usepackage[preprint]{neurips_2026}
\usepackage{soul}
\usepackage[utf8]{inputenc} 
\usepackage[T1]{fontenc}    
\usepackage{hyperref}       
\usepackage{url}            
\usepackage{booktabs}       
\usepackage{amsfonts}       
\usepackage{nicefrac}       
\usepackage{xcolor}         

\usepackage{graphicx}
\usepackage{subcaption}
\usepackage{caption}
\usepackage{multirow,multicol}
\usepackage{tabularray}
\UseTblrLibrary{booktabs}
\usepackage{bm}
\usepackage{dashrule}    
\usepackage{enumitem}

\usepackage{siunitx}
\usepackage{amsmath}
\usepackage{amssymb}
\usepackage{mathtools}
\usepackage{amsthm}
\usepackage[most]{tcolorbox}
\tcbuselibrary{breakable}
\tcbuselibrary{skins}

\definecolor{skyblue}{RGB}{128,206,255}
\definecolor{deltapos}{HTML}{1A7F37}
\definecolor{deltaneg}{HTML}{B62324}
\newcommand{\deltap}[1]{\mbox{{\color{deltapos}\textuparrow\,#1}}}
\newcommand{\deltan}[1]{\mbox{{\color{deltaneg}\textdownarrow\,#1}}}

\newtcolorbox{promptbox}[1]{
    colback=gray!10!white,
    colframe=gray!60!black,
    fonttitle=\bfseries,
    coltitle=white,
    title={#1},
    enhanced,
    attach boxed title to top left={yshift=-2mm, xshift=2mm},
    boxed title style={colback=gray!60!black},
    arc=2mm,
    drop shadow,
    left=4mm, right=4mm, top=4mm, bottom=2mm,
}

\newtcolorbox{takeaway}{
    enhanced,
    breakable,
    colback=skyblue!12,
    colframe=skyblue!75!black,
    boxrule=0.8pt,
    arc=1.5mm,
    left=4mm, right=4mm, top=3mm, bottom=2.5mm,
    fonttitle=\bfseries\small,
    coltitle=black,
    title={Takeaway},
    attach boxed title to top left={yshift=-2mm, xshift=4mm},
    boxed title style={
        colback=skyblue!30,
        colframe=skyblue!75!black,
        boxrule=0.6pt,
        arc=0.6mm,
        left=2mm, right=2mm, top=0.5mm, bottom=0.5mm,
    },
}

\usepackage{CJKutf8}
\newcommand{\cn}[1]{\begin{CJK*}{UTF8}{gbsn}#1\end{CJK*}}

\usepackage[capitalize,noabbrev]{cleveref}
\usepackage{placeins}  
\usepackage{wrapfig}   

\theoremstyle{plain}

\theoremstyle{definition}

\theoremstyle{remark}

\title{Decoding the Critique Mechanism in\\
Large Reasoning Models}

\author{%
  \parbox{.95\linewidth}{\centering
    \begin{tabular}{cccc}
      Hoang Phan$^{1}$ & Quang H. Nguyen$^{1}$ & Hung T. Q. Le$^{1}$ & Xiusi Chen$^{2}$ \\[0.6ex]
      & Heng Ji$^{2}$ & Khoa D. Doan$^{1}$\thanks{Correspondence: \texttt{khoa.dd@vinuni.edu.vn}.} &
    \end{tabular}
  }\\
  \vspace{0.8ex}\\
  $^{1}$VinUni-Illinois Smart Health Center, VinUniversity\\[0.3ex]
  $^{2}$University of Illinois Urbana-Champaign\\
}

\definecolor{hoangcolor}{HTML}{1c4ae6}

\begin{document}

\maketitle

\begin{abstract}
  Large Reasoning Models (LRMs) exhibit backtracking and self-verification mechanisms that enable them to revise intermediate steps and reach correct solutions, yielding strong performance on complex logical benchmarks.
  We hypothesize that such behaviors are beneficial only when the model has sufficiently strong ``critique'' ability to detect its own mistakes.
  This work systematically investigates how current LRMs recover from errors by inserting arithmetic mistakes in their intermediate reasoning steps.
  Notably, we discover a peculiar yet important phenomenon: despite the error propagating throughout the entire chain-of-thought (CoT) without any verbalized correction, the model still reaches the correct final answer after the thinking process finishes. 
  This recovery implies the existence of an internal mechanism helping the model to detect errors and trigger self-correction, which we refer to as the \textit{hidden critique ability}.
  Building on feature space analysis, we identify a highly interpretable \textit{critique vector} representing this behavior.
  Extensive experiments across multiple model scales and families demonstrate that steering latent representations with this vector improves the model's error detection capability and enhances the performance of test-time scaling at no extra training cost.
  Our findings provide a valuable understanding of LRMs' critique behavior, suggesting a promising direction to control and improve their self-verification mechanism.
  Our code is available at: \url{https://github.com/mail-research/lrm-critique-vectors}.
\end{abstract}

\setlength{\fboxsep}{1pt}

\section{Introduction}
The recent wave of large reasoning models (LRMs) --- including the o1 series~\cite{openai2024openaio1card}, DeepSeek-R1~\cite{Guo_2025}, and Qwen3~\cite{yang2025qwen3technicalreport} --- has raised the performance ceiling on complex, multi-step reasoning benchmarks.
Recent studies~\cite{Guo_2025, yang2025understandingahamomentsexternal} show that these models exhibit human-like ``aha moments'', including backtracking and self-verification (e.g., ``Wait, I made a mistake!''), resulting in non-linear traces rather than traditional linear CoT and enabling strong performance on complex logical benchmarks.
Yet, many ``aha moments'' and backtracking steps are redundant or illusory, leading to overthinking: models produce excessive tokens, enter self-doubt loops, and often degrade performance due to a lack of critical thinking~\cite{chen2025do, fan2025missing, zhao2025ahamomentsfakeidentifying}.
This suggests that while models possess some ``critique'' ability to detect their own mistakes, it remains intrinsically weak and not yet well understood, underscoring the need for methods that help models to reliably detect genuine errors, execute precise self-correction, and eliminate decorative reasoning steps.
Throughout this paper, we use the terms \textit{critique}, \textit{self-critique}, and \textit{self-correction} interchangeably to refer to the ability of LRMs to detect and rectify errors.

Recent work on test-time scaling increases inference compute to enhance reasoning performance, rather than scaling model size.
For instance, s1~\cite{muennighoff-etal-2025-s1} shows that minimal fine-tuning with budget forcing --- appending ``Wait'' tokens during inference --- yields significant gains, while compute-optimal strategies enable smaller models to outperform larger ones~\cite{snell2025scaling}.
However, test-time scaling does not consistently improve performance~\cite{ghosal2025does, zeng-etal-2025-revisiting}, as its effectiveness depends on the model's intrinsic self-verification capacity~\cite{setlur2025scaling, zeng-etal-2025-revisiting}.
A key bottleneck is that LLMs struggle to detect errors in long CoTs \cite{he-etal-2025-large}, fail to self-correct without external guidance \cite{kamoi-etal-2024-llms}, and cannot reliably identify their own mistakes \cite{tyen-etal-2024-llms}.
While training-based methods improve verification capabilities~\citep{weng2023large, kang2025t1toolintegratedselfverificationtesttime, chen2025sets, lee2025revise}, they require substantial training costs, architectural complexity, or external tools, reducing practicality.
We hypothesize that the budget forcing approach relies heavily on the error detection and correction ability, and investigate whether this ability can be identified and controlled without additional training.

\begin{wrapfigure}{r}{0.55\textwidth}
    \vspace{-1.0\baselineskip}
    \centering
    \includegraphics[width=\linewidth]{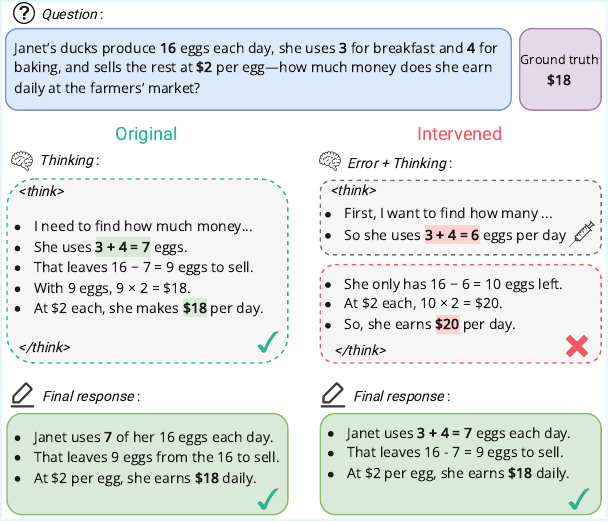}
    \caption{Hidden self-correction despite erroneous reasoning in R1-32B. Left: original correct reasoning. Right: injected error (3\,+\,4\,=\,{\colorbox{red!20}{6}}) propagates to incorrect thinking conclusion ({\colorbox{red!20}{\$20}}), yet the model recovers to the correct final answer ({\colorbox{green!20}{\$18}}). We hypothesize this indicates implicit error detection beyond the observable CoT. Full generation details are in Appendix~\ref{sec:generation_example_app}.}
    \label{fig:intervention_framework}
    \vspace{-0.5\baselineskip}
\end{wrapfigure}

Understanding how LRMs handle errors internally requires samples where models make mistakes and self-correct.
However, we observe that such instances are rare in standard generation.
When the model answers correctly, CoTs are typically all correct; when it is wrong, errors persist throughout with no clear recovery.
Moreover, labeling every step is prohibitively expensive.
Instead, we deliberately inject arithmetic errors (e.g., $3+4=6$) at intermediate steps within the reasoning trace and let the model continue without terminating its thinking process (Figure~\ref{fig:intervention_framework}).
We then compare the thinking conclusions with the ground-truth answers to assess recovery, yielding a scalable analysis framework without per-step supervision.
Surprisingly, we observe that despite the corrupted reasoning from the injected errors, the model often recovers and produces correct final answers (Figure~\ref{fig:main_observation}).
We call it the \textit{hidden critique ability}, in which LRMs silently detect errors inside \texttt{<think>...</think>} without verbalization, and rectify them after \texttt{</think>}.

To understand this recovery, we analyze the feature space and observe a strong separability between the error-recovery (with intervention) runs and the standard (without intervention) runs, although both yield correct final answers (\textit{Section}~\ref{sec:geometry}).
We hypothesize that this separability captures an internal critique mechanism, i.e., the model's hidden ability to detect and correct errors without explicit verbalization.
This hidden capability can be represented as a single linear direction in the activation space -- which is exactly the difference between the two mean vectors of activations of the intervened and standard runs. We call it the \textit{critique vector} and validate its semantic meaning with logit lens analysis across model families.
To further evaluate our hypothesis and test the utility of the critique vector, we use it to steer the LRMs and evaluate the performance on several error-detection tasks.
Experimental results show that positive steering makes LRMs more ``critical'' and improves its mistake-identification performance, while negative steering worsens it.
Finally, we show that controlling LRMs with this critique vector during test-time scaling enables the model to detect its mistakes better and produce more correct solutions.

In summary, our contributions are as follows:
\begin{itemize}[leftmargin=10pt]
\item We reveal the hidden self-correction phenomenon in LRMs that, to the best of our knowledge, has not been documented before: injected arithmetic errors corrupt CoT reasoning and intermediate conclusions, yet the model implicitly recovers to produce the correct final response.
\item Through feature space analysis, we show that activations are highly separable between the error-recovery and the standard runs. Leveraging this separation, we extract a critique vector and rigorously validate its error-detection capability through logit lens analysis and steering experiments.
\item Finally, we demonstrate that the critique vector influences test-time scaling: amplifying it improves error recognition and final accuracy, while suppressing it degrades performance.
\end{itemize}

\section{Related Work}
\textbf{Representation Engineering in LLMs.} Representation engineering extracts vectors from model activations to steer behavior without fine-tuning.
Early work~\citep{rimsky-etal-2024-steering, turner2024steeringlanguagemodelsactivation} computed steering vectors from contrastive prompt pairs, enabling applications in style control~\citep{konen-etal-2024-style} and contrastive steering~\citep{rimsky-etal-2024-steering}, later extended to refusal control~\citep{lee2025programmingrefusalconditionalactivation, arditi2024refusal} and improved instruction following~\citep{stolfo2025improvinginstructionfollowinglanguagemodels}.
Recent advances address polysemanticity through feature-level steering~\citep{yang2025lfsteeringlatentfeatureactivation} and apply these techniques to large reasoning models (LRMs), including control over thinking budgets~\citep{li2025steeringllmthinkingbudget}, trace editing~\citep{sun2025thinkeditinterpretableweightediting}, and reasoning calibration~\citep{chen2025seal}.
Unlike prior work steering style, length, or consistency, we uncover the hidden critique ability and identify a critique vector that represents latent error detection, providing mechanistic access to the internal processes behind successful test-time scaling.

\textbf{Faithfulness of Chain-of-Thought Reasoning.}
Recent studies reveal that reasoning traces in LRMs are often unfaithful to the model's actual computations --- models may fail to verbalize true reasoning~\citep{lanham2023measuring, chen2025reasoningmodelsdontsay, turpin2023language}, generate misleading explanations~\citep{arcuschin2025chainofthoughtreasoningwildfaithful}, or produce decorative ``aha moments''~\citep{zhao2025ahamomentsfakeidentifying}.
Our work observes a related phenomenon: models producing correct final answers despite corrupted reasoning.
However, we go beyond observation by locating this capability in latent space, interpreting it as a critique vector, and using it to causally control error detection and recovery.

\textbf{Error Detection and Correction.} 
Prior work reveals mixed capabilities in LLM error detection and self-correction.
While self-verification improves reasoning~\citep{weng2023large}, fundamental limitations persist: LLMs struggle to detect errors in long CoT traces~\citep{he-etal-2025-large}, cannot reliably self-correct without external guidance~\citep{kamoi-etal-2024-llms}, and only succeed when error locations are provided~\citep{tyen-etal-2024-llms}.
Recent approaches address these gaps through key condition verification~\citep{wu-etal-2024-large}, confidence--critique decomposition~\citep{yang-etal-2025-confidence}, or reinforcement learning~\citep{kumar2025training}.
In reasoning models, self-verification emerges as a critical ability during training~\citep{Guo_2025}, yet the underlying mechanisms remain underexplored.
\citet{lee2025geometryselfverificationtaskspecificreasoning} examined its geometry in task-specific models, while \citet{zhang2025reasoningmodelsknowtheyre} trained probes on reasoning steps labeled by commercial models, showing that models implicitly encode correctness information.
In contrast, we extract the model's intrinsic critique ability without external supervision, enabling mechanistic control over error detection across diverse logical reasoning tasks.

\textbf{Test-Time Scaling.} Test-time scaling shifts focus from parameter scaling to allocating more computation during inference.
Simple finetuning on reasoning traces and appending ``Wait'' tokens already yield notable gains~\citep{muennighoff-etal-2025-s1}, while compute-optimal strategies allow smaller models to outperform much larger ones~\citep{snell2025scaling}.
Follow-up work incorporates compute-optimal allocation~\citep{liu20251bllmsurpass405b}, self-verification~\citep{chen2025sets}, tool-assisted verification~\citep{kang2025t1toolintegratedselfverificationtesttime}, and reward models to assess the reasoning quality~\cite{wang2024math, zhang2024ReSTMCTS, lightman2024lets}.
However, recent studies reveal limitations to this potential, as extended thinking alone saturates and fails to scale beyond a certain point~\citep{ghosal2025does}.
Additionally, LRMs show limited scaling capabilities without proper verification~\citep{zeng-etal-2025-revisiting}, and scaling without verification or RL remains suboptimal~\citep{setlur2025scaling}.
We hypothesize self-critique to be the core mechanism underlying successful test-time scaling and show that extracted critique vectors enable direct control over this process.

\section{Uncovering Hidden Self-Correction}
\label{sec:hidden_recovery}

\subsection{Intervention Framework} \label{sec:intervention_framework}
LRMs typically operate in two phases: given the prompt $q$, they first generate a reasoning trace $c$ between a special start-of-think token $\langle\texttt{think}\rangle$ and an end-of-think token $\langle\texttt{/think}\rangle$, and then produce the final answer $a$. Formally:
\begin{equation}
    f_{\text{LRM}}: (q, \langle\texttt{think}\rangle) \mapsto (c, \langle\texttt{/think}\rangle, a).
\end{equation}

Previous work has studied the model behavior under CoT interventions via prefilling \cite{yang-etal-2025-well, wu2025effectivelycontrollingreasoningmodels}, where a variable content $e$ is integrated immediately after the thinking-start token. This generation process can be expressed as:
\begin{equation}
    f_{\text{LRM}}: (q, \langle\texttt{think}\rangle, e) \mapsto (\tilde{c}, \langle\texttt{/think}\rangle, \tilde{a}).
\end{equation}
Here, $\tilde{c}$ represents the model's reasoning continuation after intervention, and the complete CoT is formed by concatenating $e + \tilde{c}$. The intervention $e$ is flexible and task-dependent; for instance, it may be an instruction, a prompt, or another mechanism designed to analyze specific model behaviors.

A key question in our work is whether LRMs can detect and recover from errors during their internal reasoning.
To investigate this, we inject an arithmetic error $e$ (e.g., $3+4=6$) into the reasoning chain and examine whether the continuation $\tilde{c}$ recovers from it.
We generate this error using \texttt{GPT-5} with a carefully engineered prompt $\mathcal{P}_{\text{err}}$ (detailed in Appendix~\ref{app:gpt5_prompt}):
\begin{equation}
    f_{\texttt{GPT-5}}: (q, \mathcal{P}_{\text{err}}) \mapsto e.
\end{equation}

Using this approach, we construct GSM8K-Error and MATH500-Error by generating one error $e$ per question $q$, yielding a consistent benchmark across all models evaluated.
For evaluation, we extract the thinking process $\tilde{c}$ and final answer $a$, then compare both against the ground truth $y_q$ to determine whether the model can recover from the injected errors.
Unlike step-by-step supervision methods \cite{zhang2025reasoningmodelsknowtheyre} that require labeling each reasoning step, our approach is far more efficient and scalable.
Details on the error generation process and evaluation are in Appendix~\ref{sec:generation_example_app} and Section~\ref{sec:experimental_details}, respectively.

\subsection{LRMs Hiddenly Recover in the Final Answer}
\begin{wrapfigure}{r}{0.46\textwidth}
    \vspace{-1.0\baselineskip}
    \centering
    \includegraphics[width=\linewidth]{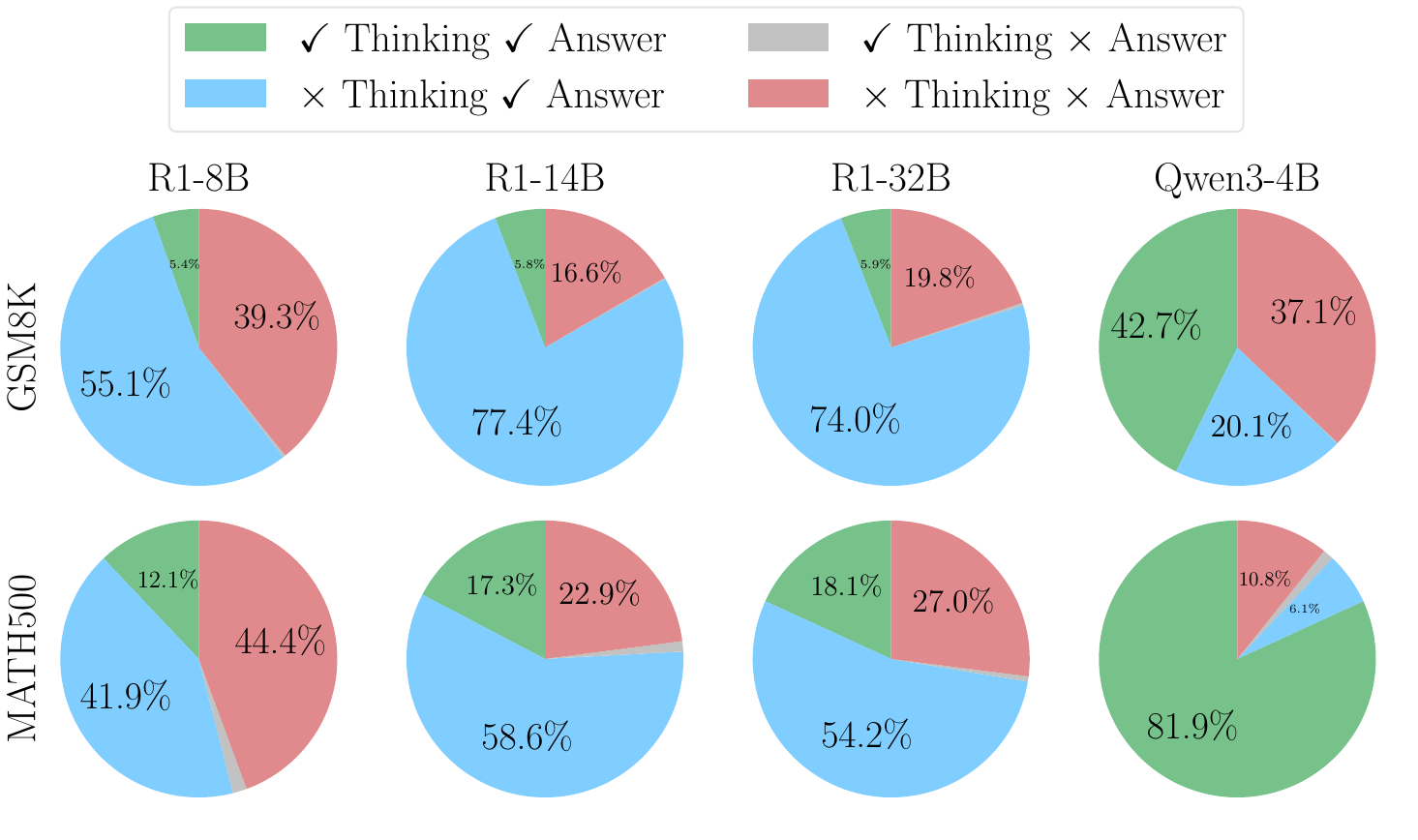}
    \caption{Distribution of reasoning-result alignment across R1 and Qwen3 model variants. The proportions represent four distinct outcomes based on the correctness of the internal ``thinking'' process versus the correctness of the final ``answer'' on the GSM8K-Error and MATH500-Error benchmarks. In this paper, we analyze the self-correction mechanisms represented by the \colorbox{skyblue}{light blue} segments: `$\times$ Think $\checkmark$ Answer'.}
    \label{fig:main_observation}
    \vspace{-0.5\baselineskip}
\end{wrapfigure}

We evaluate four open-source LRMs --- DeepSeek-R1 (8B, 14B, 32B) and Qwen3 (4B) --- on GSM8K-Error and MATH500-Error (details in Section \ref{sec:experimental_details}).
Figure~\ref{fig:main_observation} shows outcome distributions by the correctness of the internal reasoning (``Thinking'') and final output (``Answer'') on the intervened datasets.

Surprisingly, beyond the dominant cases where reasoning and final answers align (i.e., both correct or both incorrect), we observe a significant fraction where the model's visible reasoning $c$ remains corrupted yet the final answer $a$ is correct.
We hypothesize that in these cases, \textit{ LRMs implicitly correct the injected mistake without verbalizing it in the CoT}. 
We call this behavior \textit{hidden self-correction}.

Specifically, on GSM8K-Error, R1 models consistently exhibit hidden self-correction: in $55\%--78\%$ of cases, they produce correct final answers while injected errors remain intact in the reasoning trace; explicit corrections, by contrast, remain rare ($<6\%$).
This \textit{implicit recovery} persists across model scales and extends to the more challenging MATH500-Error ($41\%$--$59\%$).

Unlike R1 models, Qwen3-4B demonstrates powerful explicit backtracking, recovering from injected errors in $42.7\%$ of GSM8K-Error cases and $81.9\%$ of MATH500-Error cases.
However, a moderate fraction of samples --- $20.1\%$ on GSM8K-Error and $10.8\%$ on MATH500-Error --- still show correct final answers without CoT corrections.
This suggests that the hidden self-correction mechanism exists consistently across different model families.

Beyond our hypothesis that the model internally identifies and corrects the injected error, two alternative explanations remain:

\begin{tcolorbox}[
colback=gray!10!white,
colframe=gray!60!black,
boxrule=0.4pt,
arc=1mm,
left=0mm, right=3mm, top=2mm, bottom=2mm
]
\begin{enumerate}
    \item[\textbf{1.}] Does the model \textbf{memorize} the final answer because public benchmarks such as GSM8K or MATH500 may appear in training data?
    \item[\textbf{2.}] Does the model completely \textbf{ignore} the corrupted reasoning trace and re-solve the problem only in the final-answer space?
\end{enumerate}
\end{tcolorbox}

To test memorization, we construct a non-public synthetic GSM8K-style dataset and apply the same pipeline from Section~\ref{sec:intervention_framework}; recovery persists (Appendix~\ref{app:memorization}, Table~\ref{tab:memorization_synthetic}).
For the second case, we consider two settings: (1) masking the question so the model must rely only on the corrupted CoT, and (2) requiring the final answer to begin with ``Based on my thinking above'' to encourage reliance on the trace.
The $\times$ Think / $\checkmark$ Answer pattern persists in both cases (Appendix~\ref{app:error_injected_variants}, Figures~\ref{fig:variant_qmask}--\ref{fig:variant_ansprefix}, Table~\ref{tab:error_injected_variants_app}).

Overall, these results confirm that the phenomenon reflects a hidden self-correction mechanism, not an artifact of memorization or trace-bypassing.

\section{The Geometry of Critique Ability}
\label{sec:geometry}

Section~\ref{sec:hidden_recovery} shows that LRMs possess a hidden critique ability to recover from injected errors, yet this behavior almost disappears on natural CoTs (Table~\ref{tab:hidden_critique_original_samples}).
In this section, we compare these two cases to extract and analyze the critique ability using mechanistic interpretability tools.
We observe linear separability in the latent space between the intervened (incorrect-CoT) and baseline (correct-CoT) runs (Section~\ref{sec:activation_separability}).
This linear separability suggests that the hidden capability can be decoded as a single direction in the activation space.
We validate this hypothesis via Logit Lens \cite{nostalgebraist2020logitlens} and extract a steering vector that capture the recovery mechanism.

\begin{table}[t]
\centering
\caption{Distribution of reasoning and final answer correctness on GSM8K and MATH500.}
\label{tab:hidden_critique_original_samples}
\footnotesize
\setlength{\tabcolsep}{4pt}
\begin{tabular}{@{}l
S[table-format=2.1, table-space-text-post=\%]
S[table-format=2.1, table-space-text-post=\%]
S[table-format=2.1, table-space-text-post=\%]
S[table-format=2.1, table-space-text-post=\%]
S[table-format=2.1, table-space-text-post=\%]
S[table-format=2.1, table-space-text-post=\%]
S[table-format=2.1, table-space-text-post=\%]
S[table-format=2.1, table-space-text-post=\%]
@{}}
\toprule
 \multirow{2}{*}{Cases}& \multicolumn{4}{c}{GSM8K} & \multicolumn{4}{c}{MATH500} \\ \cmidrule(lr){2-5}\cmidrule(lr){6-9}
& \multicolumn{1}{c}{Q3-4B}
& \multicolumn{1}{c}{R1-8B}
& \multicolumn{1}{c}{R1-14B}
& \multicolumn{1}{c}{R1-32B}
& \multicolumn{1}{c}{Q3-4B}
& \multicolumn{1}{c}{R1-8B}
& \multicolumn{1}{c}{R1-14B}
& \multicolumn{1}{c}{R1-32B} \\ \midrule
$\times$ Think, $\times$ Ans & 3.9\% & 15.6\% & 6.8\% & 5.0\% & 1.5\% & 5.0\% & 4.1\% & 2.4\% \\
$\checkmark$ Think, $\checkmark$ Ans & 95.8\% & 81.8\% & 92.4\% & 94.0\% & 98.3\% & 95.0\% & 95.5\% & 97.4\% \\
$\checkmark$ Think, $\times$ Ans & 0.0\% & 0.7\% & 0.2\% & 0.5\% & 0.2\% & 0.0\% & 0.0\% & 0.0\% \\
$\times$ Think, $\checkmark$ Ans & 0.2\% & 1.9\% & 0.5\% & 0.5\% & 0.0\% & 0.0\% & 0.4\% & 0.2\% \\ \bottomrule
\end{tabular}
\end{table}

\subsection{The Behavior on Original Samples}\label{sec:hidden_critique_original_samples}
Section~\ref{sec:hidden_recovery} shows that LRMs secretly detect \textit{injected errors} and recover in the final answer. 
Here, we investigate the behavior of LRMs on original samples by measuring the accuracy of the thinking and the final answer separately.
Table~\ref{tab:hidden_critique_original_samples} demonstrates that CoT and final answer are usually consistent: when the thinking is wrong, the model hardly fixes it and thus returns the incorrect final answer.
The recovery rate is only $1.9\%$ for R1-8B on GSM8K and less than $1\%$ in every other case.
These results suggest that, without injected errors, LRMs do not invoke critique behavior: correct reasoning and correct answers typically co-occur, while recovery from faulty CoTs is rare.

\subsection{The Linear Separability of Internal Thoughts}
\label{sec:activation_separability}

\begin{wraptable}[14]{r}{0.48\textwidth}
    \vspace{-1.0\baselineskip}
    \centering
    \footnotesize
    \caption{Linear probing results. Comparison of AUROC and ECE evaluated on GSM8K (ID) and MATH500 (OOD) across four different models. The reported results are selected for the best layer. Detailed results can be found in Appendix~\ref{sec:activation_separability_app}.}
    \label{tab:linear_probing_results_main}
    \begin{tblr}{
        width = \linewidth,
        colspec = {X[l,m,1.25] X[c,m,0.9] X[c,m,0.8] X[c,m,0.9] X[c,m,0.8] X[c,m,0.85]},
        colsep = 3.0pt,
        rowsep = 1.2pt,
        rows = {ht = 1.55em},
    }
        \toprule
        \SetCell[r=2]{c} Model 
        & \SetCell[c=2]{c} GSM8K & 
        & \SetCell[c=2]{c} MATH500 & 
        & \SetCell[r=2]{c} Layer \\
        \cmidrule[lr]{2-3} \cmidrule[lr]{4-5}
        & AUROC & ECE & AUROC & ECE & \\
        \midrule
        Qwen3-4B  & 1.000 & 0.002 & 1.000 & 0.026 & 21 \\
        R1-8B  & 0.998 & 0.019 & 0.996 & 0.022 & 31 \\
        R1-14B & 1.000 & 0.003 & 0.998 & 0.012 & 28 \\
        R1-32B & 1.000 & 0.002 & 0.997 & 0.008 & 38 \\
        \bottomrule
     \end{tblr}
    \vspace{-0.5\baselineskip}
\end{wraptable}

We evaluate on in-distribution (ID) GSM8K-Error and out-of-distribution (OOD) MATH500-Error,
using AUROC and Expected Calibration Error (ECE).
Table~\ref{tab:linear_probing_results_main} reports results for the best layer per model (highest ID AUROC, lowest ID ECE), with full layer-wise results in Appendix~\ref{sec:activation_separability_app}.

To further investigate, we compare the latent features of intervened samples in Section~\ref{sec:hidden_recovery} and original samples.
We write $x \sim y_q$ when an output $x$, which could be a reasoning conclusion $c$ or a final answer $a$, matches the ground-truth $y_q$, and $x \not\sim y_q$ otherwise.
For 1,000 GSM8K training questions $\mathcal{Q}_{\text{train}}$, we construct paired runs with identical prompts: a baseline run with correct reasoning ($c \sim y_q$) and an intervened run with incorrect reasoning ($e + \tilde{c}  \not\sim  y_q$), both of which yield correct final answers ($a \sim y_q, \tilde{a} \sim y_q$).
As discussed above, in the final answer the model performs hidden critique ability to recover from mistakes.
Therefore, we inspect the hidden states of the final answers and extract the average feature vectors across answer tokens at each layer $\ell$:
\begin{equation}\label{eq:activation_extraction}
\begin{aligned}
h_{\text{baseline}}^{(\ell)}(q) &= \frac{1}{|a|} \sum_{i \in a} h_i^{(\ell)}(q, c, a), \quad
h_{\text{intervened}}^{(\ell)}(q) &= \frac{1}{|\tilde{a}|} \sum_{i \in \tilde{a}} h_i^{(\ell)}(q, e + \tilde{c}, \tilde a),
\end{aligned}
\end{equation}
where $c \sim y_q,\; a \sim y_q,\;e + \tilde{c} \not \sim y_q, \; \tilde{a} \sim y_q$.

For each layer $\ell$, we train a linear probe $f_{\ell}(h)=W_{\ell}h+b_{\ell}$ with $W_{\ell}\in\mathbb{R}^{2\times d}$, where $d$ is the size of feature vectors, to distinguish the baseline from intervened activations.
The probe is trained with AdamW (learning rate $10^{-3}$) for 100 epochs, with a cosine schedule.

We observe consistently strong probing performance across all models and both datasets, with near-perfect AUROC.
This indicates that the underlying activations remain linearly separable despite the matching final answers.
This separability suggests that the hidden self-correction ability discovered in Section \ref{sec:hidden_recovery} corresponds to a single linear direction in the latent space.

One concern is whether this near-perfect separability reflects \textbf{hidden self-correction} or an activation shift induced by the \textbf{corrupted reasoning style}. 
Appendix~\ref{sec:probe_validity_app} rules this out: recovered and unrecovered samples --- despite sharing the same corrupted CoT --- remain representationally distinct.

\subsection{Extracting the Critique Vector}\label{sec:vector_extraction}
Following previous works~\cite{turner2024steeringlanguagemodelsactivation, rimsky-etal-2024-steering}, we identify the \textit{critique vector} as the difference-in-means between contrastive pairs of runs constructed in Eqn~\ref{eq:activation_extraction}.
We consider the intervened runs in which the model performs hidden self-correction as positive examples, and the baseline runs in which the model merely summarizes the CoTs without invoking recovery as negative examples.

Formally, we compute the critique vector at layer $\ell$ as the mean difference between intervened and baseline activations over the training dataset $\mathcal{Q}_{\text{train}}$:
\begin{equation}
v_{\ell} = \frac{1}{|\mathcal{Q}_{\text{train}}|} \sum_{q \in \mathcal{Q}_{\text{train}}} \Bigl( h_{\text{intervened}}^{(\ell)}(q) - h_{\text{baseline}}^{(\ell)}(q) \Bigr).
\label{eq:steering_vector}
\end{equation}

\subsection{Steering with the Critique Vector}

\begin{wraptable}{r}{0.48\textwidth}
    \vspace{-1.0\baselineskip}
    \centering
    \scriptsize
    \caption{Comparison of the top 10 predicted tokens and their associated logit scores for the Qwen3-4B and R1-32B models.}
    \label{tab:logit_lens_main}
    \begin{tblr}{
        width = \linewidth,
        colspec = {
            Q[c,m,2.0em]
            X[l,m,1.15]
            Q[c,m,2.8em]
            X[l,m,1.65]
            Q[c,m,2.8em]
        },
        colsep = 2pt,
        rowsep = 0.5pt,
        row{1-2} = {bg=gray!25},
        row{4,6,8,10,12} = {bg=gray!10},
        rows = {ht = 1.5em}
    }
        \toprule
        \SetCell[r=2]{c} \textbf{Rank}
        & \SetCell[c=2]{c} \textbf{Qwen3-4B}
        & & \SetCell[c=2]{c} \textbf{R1-32B} & \\
        \cmidrule[lr]{2-3} \cmidrule[lr]{4-5}
        & \textbf{Token} & \textbf{Logit} & \textbf{Token} & \textbf{Logit} \\
        \midrule
        1  & wrongly   & 1.78 & Wait & 2.68 \\
        2  & Nope      & 1.72 & {\cn{但实际上} \scriptsize (But actually)} & 2.65 \\
        3  & {\cn{而且还} \scriptsize (And also)} & 1.63 & {\cn{却} \scriptsize (But/Yet)} & 2.64 \\
        4  & really    & 1.61 & however & 2.60 \\
        5  & \_that    & 1.60 & but & 2.58 \\
        6  & Actually  & 1.60 & {\cn{不过} \scriptsize (However)} & 2.57 \\
        7  & {\cn{但却} \scriptsize (But yet)} & 1.59 & Wait & 2.55 \\
        8  & ?)\textbackslash n\textbackslash n & 1.57 & ( & 2.49 \\
        9  & oque      & 1.57 & wait & 2.41 \\
        10 & actually  & 1.56 & However & 2.41 \\
        \bottomrule
    \end{tblr}

\end{wraptable}

We can use the critique vector in Eqn~\ref{eq:steering_vector} to causally control and modulate the model's hidden critique/recovery ability.
Following the standard steering approach~\cite{venhoff2025understanding, sun2025thinkeditinterpretableweightediting, arditi2024refusal, chen2025seal, yang2025lfsteeringlatentfeatureactivation}, we add a scaled steering vector to the residual stream immediately after the MLP layer:
\begin{equation}
h_{\text{steered}}^{(\ell)} = h^{(\ell)} + \alpha \, v_{\ell},
\label{eq:steering_vector_addition}
\end{equation}
where $h^{(\ell)}$ is the post-MLP residual stream activation at layer $\ell$, $v_{\ell}$ is the steering vector at layer $\ell$, and $\alpha \in \mathbb{R}$ is a scalar steering coefficient.
The modified activations $h_{\text{steered}}^{(\ell)}$ are then propagated through subsequent layers, allowing the steering direction to shape the model's downstream generation.
Intuitively, this steering vector represents the \textit{critique ability} of LRMs, which helps them detect and recover from the injected mistakes.
Positive values of $\alpha$ amplify the model's self-critique ability, whereas negative values suppress this behavior. Further details on how we select the layer $\ell$ for steering are provided in Appendix \ref{app:layer_effect}.

\subsection{Logit Lens Interpretation}
In this section, we investigate the meaning of $v_{\ell}$ using Logit Lens~\cite{nostalgebraist2020logitlens}, which estimates the model's belief of its internal states.
Specifically, we project $v_{\ell}$ into the vocabulary space using the unembedding matrix $W_U \in \mathbb{R}^{d \times |V|}$, where $|V|$ denotes the size of the vocabulary:
\begin{equation}
\text{logits}_{\ell} = v_{\ell} \, W_U^\top \in \mathbb{R}^{|V|}.
\label{eq:logit_lens_steering}
\end{equation}

We interpret $v_{\ell}$ by projecting it onto the vocabulary and extracting the highest-ranked tokens.
Table~\ref{tab:logit_lens_main} lists the top 10 tokens for Qwen3-4B (layer 29) and R1-32B (layer 57).
Notably, tokens such as \textit{`Nope'} or \textit{`Wait'} rank highly --- terms typically associated with reasoning reassessment.
Both models also show strong activation of Chinese adversative markers, including \cn{但却}' (but yet), \cn{但实际上}' (but actually), and `\cn{却}' (but/yet), suggesting that $v_{\ell}$ encodes critique behavior across languages.
Additional results for multiple layers can be found in Appendix \ref{app:logit_lens}.

\FloatBarrier
\section{Experiments}

In this section, we confirm the role of the critique vector found in Section~\ref{sec:vector_extraction}, which allows LRMs to \textit{detect} errors in CoTs and \textit{correct} them in the final answer.
We causally intervene on the feature space with the critique vector to evaluate the detection ability in Section~\ref{sec:error_detection}.
Furthermore, we demonstrate the importance of the correction ability for test-time scaling in Section~\ref{sec:error_recovery}.

\subsection{Experimental Details} \label{sec:experimental_details}

\textbf{Datasets.} We intervene on two widely-used mathematics benchmarks, GSM8K~\cite{cobbe2021trainingverifierssolvemath} and MATH500~\cite{hendrycks2021measuringmathematicalproblemsolving}, to observe model behavior under injected errors.
The GSM8K training split is used to train linear probes and derive steering vectors.
For error detection evaluation, we use two human-annotated benchmarks: ProcessBench~\cite{zheng-etal-2025-processbench} for Olympiad-level mathematics and BIG-Bench-Mistake~\cite{tyen-etal-2024-llms} for diverse logical reasoning tasks.
For test-time scaling and self-correction analysis, we use our constructed GSM8K-Error and MATH500-Error datasets alongside BIG-Bench-Mistake, which supports correction from erroneous chains-of-thought.

\textbf{Models.} 
We evaluate two open-source reasoning model families: Qwen3~\cite{yang2025qwen3technicalreport} and DeepSeek-R1~\cite{Guo_2025} across different scales: \href{https://huggingface.co/Qwen/Qwen3-4B}{Qwen3-4B}, \href{https://huggingface.co/deepseek-ai/DeepSeek-R1-Distill-Llama-8B}{DeepSeek-R1-Distill-Llama-8B}, \href{https://huggingface.co/deepseek-ai/DeepSeek-R1-Distill-Qwen-14B}{DeepSeek-R1-Distill-Qwen-14B}, and \href{https://huggingface.co/deepseek-ai/DeepSeek-R1-Distill-Qwen-32B}{DeepSeek-R1-Distill-Qwen-32B}, covering both Llama and Qwen architectures within R1, hereafter Qwen3-4B, R1-8B, R1-14B, and R1-32B.

\begin{figure}[!tp]
    \centering
    \begin{subfigure}[t]{0.49\linewidth}
        \centering
        \includegraphics[width=\linewidth]{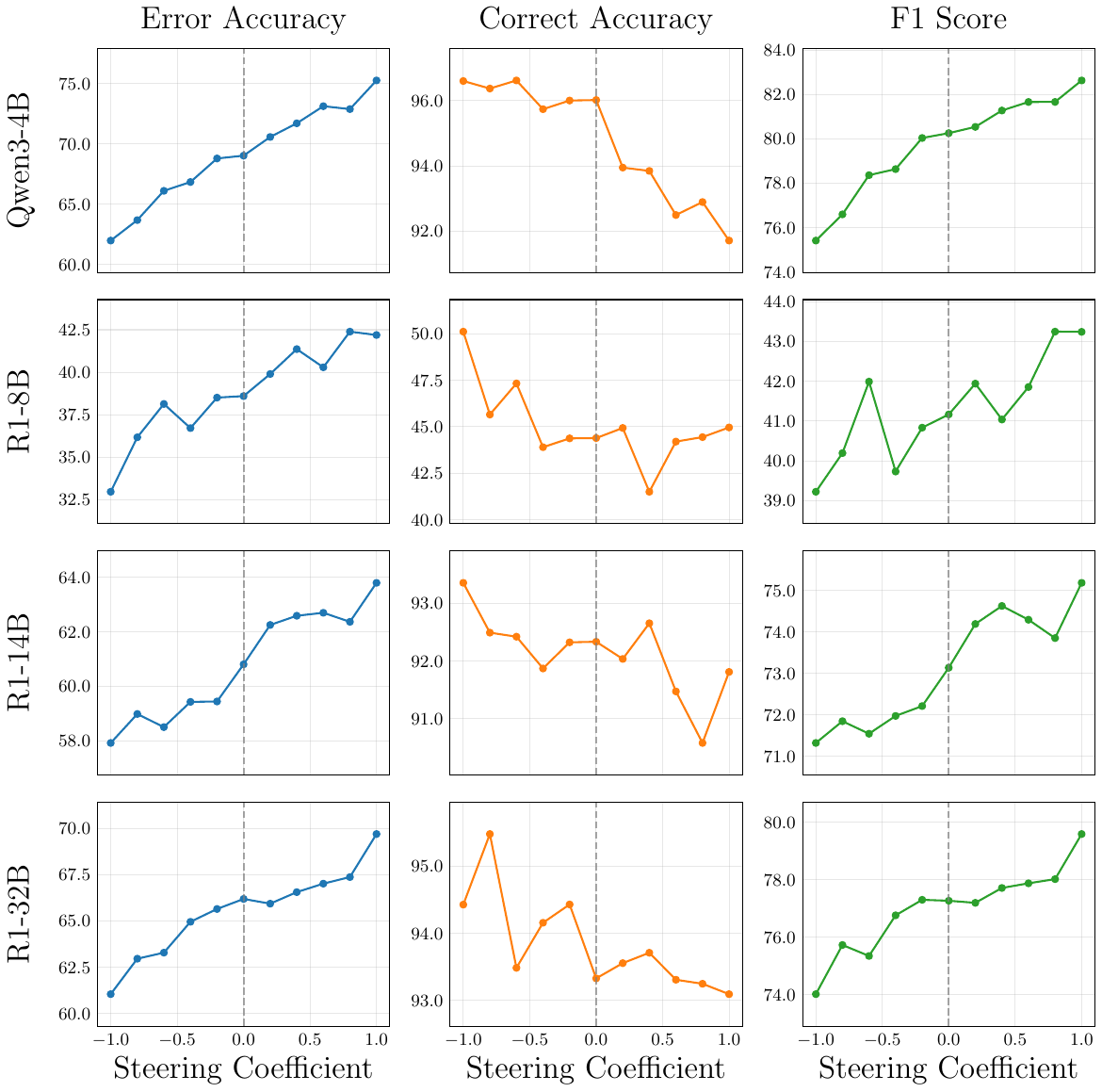}
        \caption{ProcessBench.}
        \label{fig:process_bench_results}
    \end{subfigure}
    \hfill
    \begin{subfigure}[t]{0.49\linewidth}
        \centering
        \includegraphics[width=\linewidth]{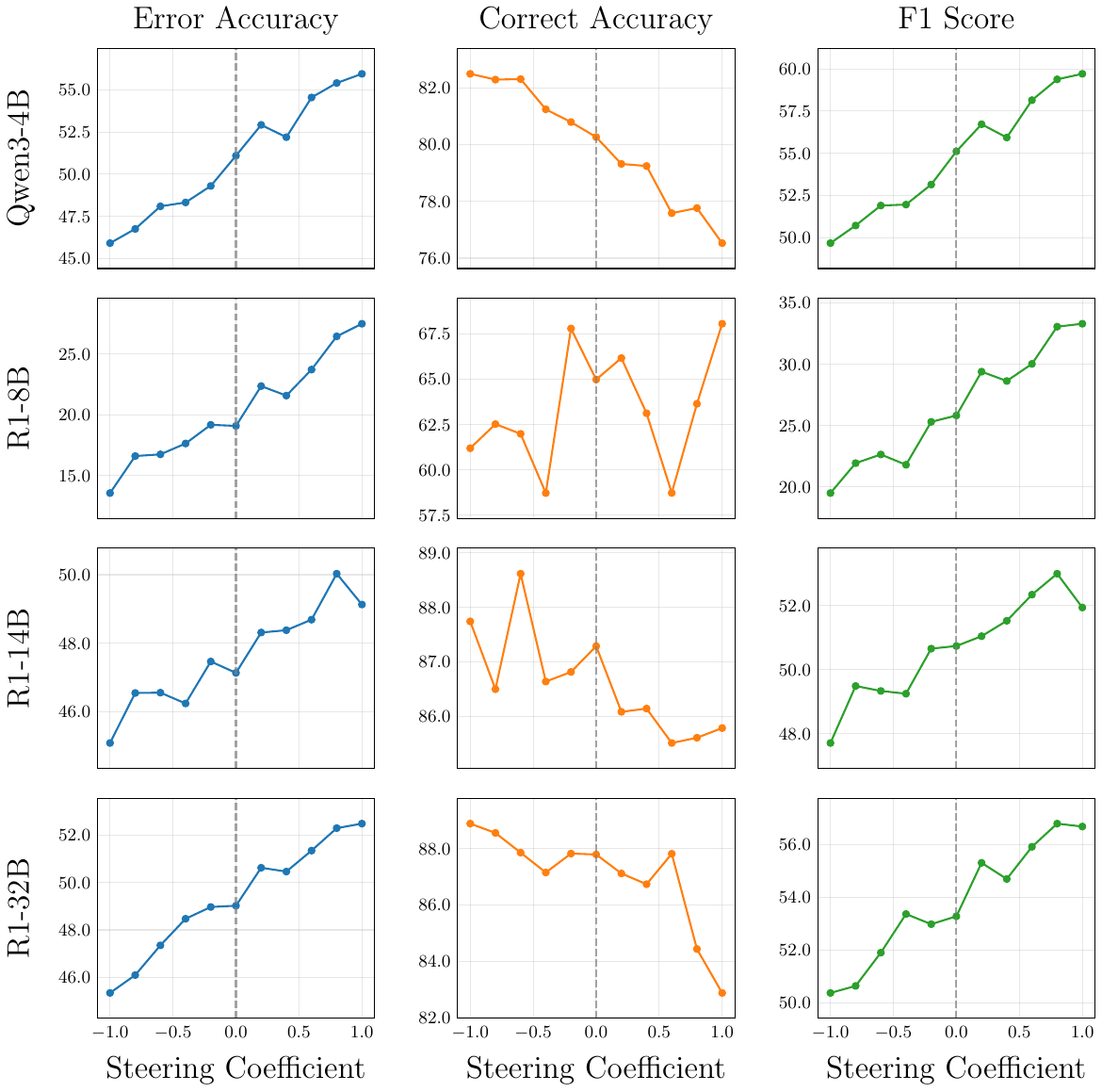}
        \caption{BIG-Bench Mistake.}
        \label{fig:bigbench_results}
    \end{subfigure}
    \caption{Effect of steering coefficient $\alpha$ on error detection accuracy, correct solution accuracy, and F1 score. The monotonic relationship between $\alpha$ and error detection holds on both ProcessBench and BIG-Bench Mistake. The dashed vertical line indicates the baseline performance at $\alpha=0$.}
    \label{fig:error_detection_results}
\end{figure}

\textbf{Evaluation.} 
We use \href{https://huggingface.co/IAAR-Shanghai/xVerify-0.5B-I}{xVerify-0.5B-I} to verify both reasoning traces and final answers, as it provides state-of-the-art math evaluation without requiring human annotation and avoids template mismatch~\citep{chen2025xverifyefficientanswerverifier}.
For feature analysis, we use greedy decoding (temperature 0.0) to study deterministic internal behavior.
For steering experiments, we set the temperature to 0.6 following official guidelines to prevent reasoning loops --- repetitive thinking patterns common in reasoning models, which we also observed with greedy decoding on challenging tasks~\citep{pipis2025waitwaitwaitreasoning}.
We fix all random seeds to 0 to ensure complete reproducibility across runs.
All experiments use two H100 GPUs.


\definecolor{bg_problem}{RGB}{240, 245, 255}
\definecolor{bg_negative}{RGB}{255, 245, 245}
\definecolor{bg_positive}{RGB}{245, 255, 245}
\definecolor{bg_baseline}{RGB}{250, 250, 250}

\definecolor{border_problem}{RGB}{50, 50, 150}
\definecolor{border_positive}{RGB}{0, 100, 0}
\definecolor{border_negative}{RGB}{150, 0, 0}
\definecolor{border_baseline}{RGB}{100, 100, 100}

\definecolor{txt_correct}{RGB}{0, 120, 0}
\definecolor{txt_error}{RGB}{200, 0, 0}

\tcbset{
    paperbox/.style={
        enhanced,
        fonttitle=\bfseries\sffamily\small,
        fontupper=\fontsize{6pt}{7pt}\selectfont\sffamily,
        fontlower=\fontsize{6pt}{7pt}\selectfont\sffamily,
        boxrule=0.4pt,
        left=0.4mm, right=0.4mm, top=0.3mm, bottom=0.3mm,
        before skip=1pt,
        after skip=1pt,
        arc=2pt,
        segmentation style={solid, black!50!white},
    }
}

\subsection{Steering for Error Detection}\label{sec:error_detection}

\textbf{Setup.} 
We conduct experiments on ProcessBench and BIG-Bench Mistake, where each sample contains a question and a CoT with at most one erroneous step.
LRMs are prompted to return the index of the incorrect step, or $-1$ if no mistake exists (details in Appendix~\ref{app:error_detect_prompt}).
Following the original evaluation protocol of each dataset, we report Error Accuracy (CoTs with mistakes), Correct Accuracy (CoTs without mistakes), and F1 Score over all samples.

\begin{figure}[t]

    \input{qualitative_examples/main_ex_pos}
    \caption{Qualitative example of positive steering on BIG-Bench arithmetic task with R1-32B. There is a subtle arithmetic error introduced in the prompt ($7+2+20=30$, whereas it should be $29$). While the baseline model hallucinates that the math is correct during verification, positive steering enables the model to successfully catch the addition error.}
\label{fig:qual_ex_main}
\vspace{-1.8\baselineskip}
\end{figure}

\begin{wrapfigure}{r}{0.52\textwidth}
    \vspace{-1.0\baselineskip}
    \centering
    \includegraphics[width=\linewidth]{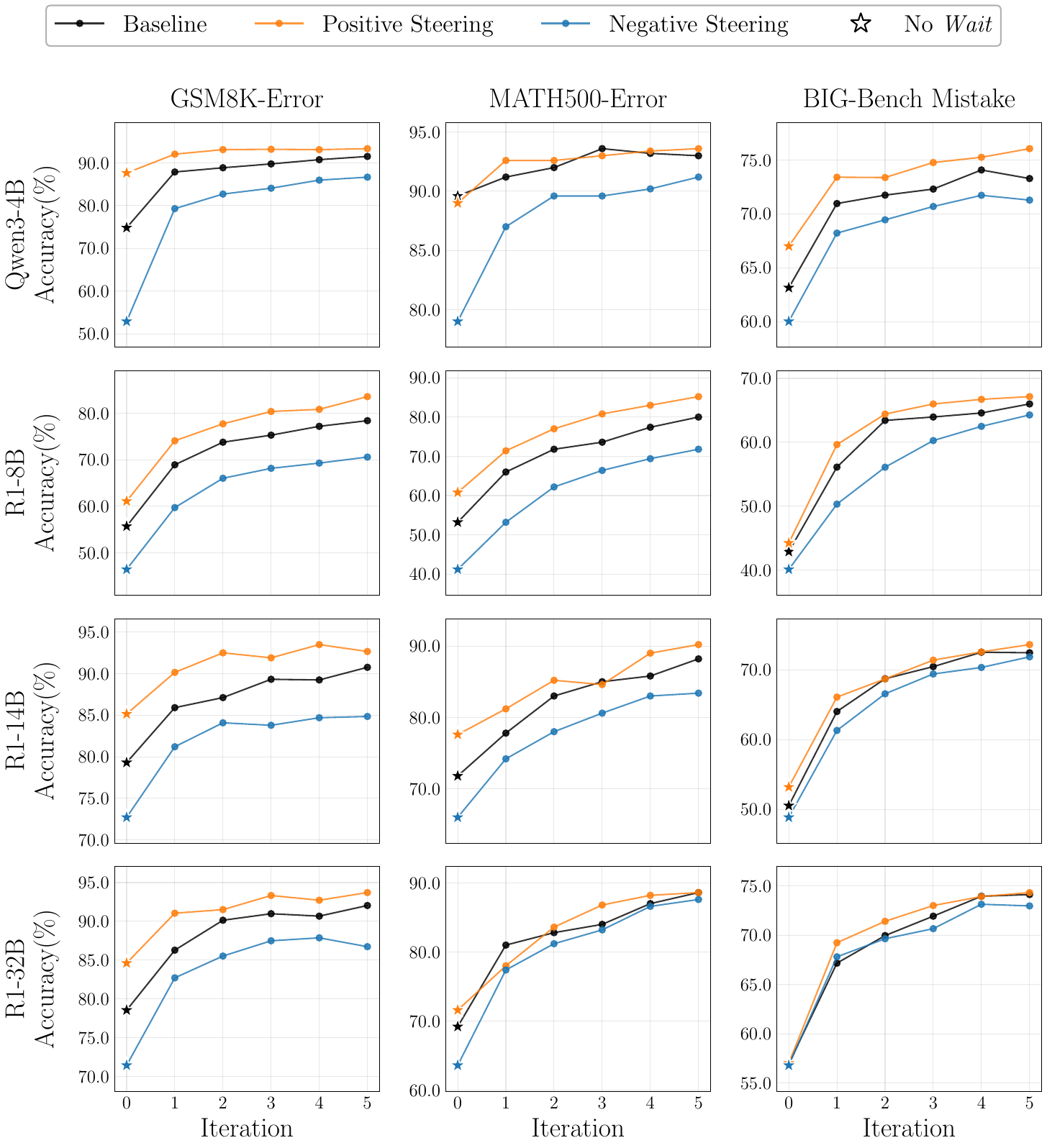}
    \caption{Impact of steering on model accuracy for test-time scaling. \textbf{Left to right:} Accuracy across three benchmarks: GSM8K-Error, MATH500-Error, and BIG-Bench Mistake. For increasing iterations of test-time scaling, positive steering consistently enhances the performance of all models.}
    \label{fig:tts_results}
    \vspace{-1.2\baselineskip}
\end{wrapfigure}
\textbf{Critique vector controls the detection ability.} 
We steer LRMs with the critique vector using coefficients ranging from $-1.0$ to $1.0$ (Figures \ref{fig:process_bench_results} and \ref{fig:bigbench_results}).
Overall, there is a positive correlation between the steering coefficients and model performance, confirming that the critique vector causally controls error detection.
On ProcessBench, positive steering $(\alpha=1.0)$ improves error detection accuracy by $5\%$ on all models. In contrast, negative steering consistently decreases it.
Similarly, on BIG-Bench Mistake, positive steering raises Qwen3-4B's error accuracy by 4\% (from $52\%$ to $56\%$).
Notably, Qwen3-4B outperforms R1-distilled models across both datasets.
A positive-steering example is shown in Figure~\ref{fig:qual_ex_main}; corresponding negative-steering examples are deferred to Appendix~\ref{sec:error_qualitative_app}.

We notice that in few scenarios, positive steering could harm the correct accuracy, such as Qwen3-4B; qualitative examples of this failure mode are provided in Appendix~\ref{sec:error_qualitative_app}.
We hypothesize that this intervention forces the model to revise the solution, potentially misunderstanding the correct thought and inadvertently recognizing it as incorrect.
However, the drop in correct accuracy is insignificant, leaving the overall positive correlation between F1 Score and steering coefficient intact.


\subsection{Test-Time Scaling with Critique Steering}
\label{sec:error_recovery}

\textbf{Setup.} 
We evaluate error recovery performance on GSM8K-Error, MATH500-Error, and BIG-Bench Mistake. 
For GSM8K-Error and MATH500-Error, the model generates from the injected error until it finishes the final answer (details of test-time scaling instruction prompt in Appendix~\ref{app:tts_prompt}).
We compare the final answer with the ground-truth label to compute the accuracy.
To test if test-time scaling helps fix injected errors, the budget forcing~\cite{muennighoff-etal-2025-s1} is applied by appending ``Wait'' token at each thinking termination attempt.
We report the performance at different numbers of iterations corresponding to the number of appended ``Wait'' tokens.
For the critique vector effect, we perform positive steering with a coefficient $1.0$ and negative steering with a coefficient $-1.0$.

\textbf{Critique vector steers the recovery performance.} 
Figure~\ref{fig:tts_results} depicts that steering with the critique vector successfully controls the error recovery ability.
On GSM8K-Error, Qwen3-4B achieves $77\%$ accuracy at baseline; positive steering boosts this to nearly $90\%$, while negative steering degrades the performance to slightly above $50\%$.
Similar trends hold across other R1-distilled models and datasets (MATH500-Error and BIG-Bench Mistake).
This confirms that error correction is causally driven by the critique ability encoded by our vector $v_\ell$ in Section~\ref{sec:vector_extraction}.

\textbf{Test-time scaling corrects the injected mistake.} 
Besides the hidden critique ability, explicit reflection via budget forcing also recovers from injected errors.
Indeed, Figure~\ref{fig:tts_results} shows that appending ``Wait'' reliably improves the accuracy across all models and datasets, aligning with prior findings on test-time scaling~\cite{muennighoff-etal-2025-s1, snell2025scaling}.

\textbf{Critique vector further enhance test-time scaling.}
However, test-time scaling alone does not fully repair the incorrect CoT: positive steering consistently boosts accuracy across all forcing iterations, while negative steering reduces it.
These results indicate that hidden critique is complementary to and can not be replaced by explicit backtracking.

\subsection{Steering on Standard Reasoning Benchmarks}
\label{sec:standard_steering}

We further apply the critique vector to original unperturbed benchmarks (Table~\ref{tab:tts_standard}). 
Across all four models, positive steering yields consistent gains and negative steering hurts across all models, confirming that the critique direction transfers beyond the injected-error setting.

We hypothesize that there could be two reasons for the modest gains.
First, the derivation data is simple: a vector trained on GSM8K-Error captures arithmetic-style critique and transfers weakly to AIME, where errors reflect missing knowledge rather than failed self-checks. 
Second, the steering is \emph{unconditional}: a fixed vector is added at every step, which on near-saturated tasks adds little signal and can over-revise correct traces (cf.\ Section~\ref{sec:error_detection}).
Both are scalable engineering problems that we leave to future work.
More broadly, our critique vector offers a concrete representational target for \emph{training} better verifiers --- for example, as a supervision signal for representation-level finetuning~\citep{wu2024reft, zou2025representationengineeringtopdownapproach} --- complementing its inference-time use.

\begin{table}[!htbp]
\centering
\scriptsize
\vspace{-0.25\baselineskip}
\caption{Steering accuracy (\%) on standard (unperturbed) reasoning benchmarks, using the same GSM8K-Error-derived vector as Figure~\ref{fig:tts_results}. Columns under each dataset correspond to steering coefficient $\alpha\in\{-1,0,+1\}$; \textbf{bold} marks the best value within each row's dataset block.}
\label{tab:tts_standard}
\small
\setlength{\tabcolsep}{4pt}
\begin{tabular*}{\linewidth}{@{\extracolsep{\fill}}l ccc ccc ccc ccc @{}}
\toprule
Model & \multicolumn{3}{c}{AIME24} & \multicolumn{3}{c}{AIME25} & \multicolumn{3}{c}{GSM8K} & \multicolumn{3}{c}{MATH500} \\
\cmidrule(lr){2-4} \cmidrule(lr){5-7} \cmidrule(lr){8-10} \cmidrule(lr){11-13}
& $-1$ & $0$ & $+1$ & $-1$ & $0$ & $+1$ & $-1$ & $0$ & $+1$ & $-1$ & $0$ & $+1$ \\
\midrule
Qwen3-4B & 50.0 & \textbf{56.7} & \textbf{56.7} & 50.0 & 46.7 & \textbf{53.3} & 50.8 & 57.7 & \textbf{62.9} & 92.2 & \textbf{93.0} & 92.8 \\
R1-8B    & 40.0 & \textbf{50.0} & \textbf{50.0} & 30.0 & 30.0 & 30.0 & 69.4 & \textbf{70.7} & \textbf{70.7} & 89.2 & \textbf{89.9} & \textbf{89.9} \\
R1-14B   & 56.7 & 56.7 & \textbf{60.0} & 36.7 & \textbf{46.7} & \textbf{46.7} & 68.6 & \textbf{71.3} & 70.3 & 92.8 & 94.2 & \textbf{95.6} \\
R1-32B   & 60.0 & 63.3 & \textbf{66.7} & 50.0 & \textbf{53.3} & \textbf{53.3} & 89.1 & 89.2 & \textbf{89.6} & 94.2 & 94.3 & \textbf{95.0} \\
\bottomrule
\end{tabular*}
\vspace{-0.5\baselineskip}
\end{table}

\FloatBarrier
\section{Conclusion}

In this paper, we identify a \emph{hidden critique behavior} in LRMs: they silently recover from injected mistakes in CoTs and produce correct final answers without verbalizing the error-detection or correction process.
Through feature analysis, we show that LRMs internally separate intervened samples that exhibit this behavior from original ones that do not, and we extract a \emph{critique vector} representing this mechanism, validated via logit lens.
We identify the critique vector representing this hidden mechanism and validate it through logit lens analysis. 
Empirically, steering with this vector successfully controls error detection across several benchmarks and models. 
We also show that test-time scaling relies on critique ability and can be combined with critique steering to further enhance correction performance. 
Our findings shed light on the role of critique ability in current LRMs and open many interesting questions. Future work should investigate whether this behavior stems from pre- or post-training and how it emerges. Since hidden critique ability leads LRMs to silently correct errors without verbalizing them in CoTs, its impact on CoT faithfulness, alignment, and other properties warrants further study. Finally, leveraging this ability to improve test-time scaling methods such as verifier-based approaches, self-consistency, and self-correction prompting remains an exciting direction we leave to future work.

\bibliographystyle{unsrtnat}
\bibliography{main}

\newpage
\appendix

\section{System Prompt}
\label{app:prompts}

\subsection{GPT-5 Instruction Prompt}
\label{app:gpt5_prompt}

The following system prompt $\mathcal{P}_{\text{err}}$ was used to generate incorrect reasoning steps:

\begin{promptbox}{GPT-5 Instruction Prompt}
Given a question and its ground-truth answer, generate a natural reasoning trace that contains a calculation error.

\medskip
\textbf{PURPOSE:} \\
This trace is a deliberate error variant used to test a reasoning model's ability to recover from arithmetic mistakes.

\medskip
\textbf{INSTRUCTIONS:} \\
- Write the trace as a sequence of short, natural reasoning steps, phrased as if someone is thinking aloud. \\
- Use natural, anthropomorphic openings such as ``First, I need to...'', ``Okay, so...'', ``Let me...'', or other common starter phrases typically found in the thinking process of reasoning models. \\
- Separate steps with two newlines ($\backslash$n$\backslash$n). \\
- Do NOT mention or hint that an error occurred (no phrases like ``mistake'', ``should be'', or ``wrong''). \\
- Introduce an arithmetic error at the earliest plausible calculation step. \\
- Stop the reasoning immediately after the mistaken calculation step; do not continue solving or produce an answer. The goal is to isolate the reasoning up to the error only. \\
- The trace must be self-consistent and match the question up to the step before the error. \\
- Keep the reasoning coherent, natural, and plausible.

\medskip
\textbf{EXAMPLE:} \\
Question: Olivia buys 3 packs of markers, each pack containing 12 markers. She gives 7 markers to her friend. How many markers does Olivia have left? \\
Ground-truth answer: 29

\medskip
Trace: ``Okay, so I need to figure out how many markers Olivia got in total. She bought 3 packs, each with 12 markers.$\backslash$n$\backslash$nNext, I'll multiply $3 \times 12 = 38$.''

\medskip
Return only the interventions using the configured JSON schema.

\medskip
\textbf{Question:} \\
\textcolor{cyan}{\{question\}} \\
\textbf{Ground-truth answer:} \\
\textcolor{cyan}{\{ground\_truth\}}
\end{promptbox}

\subsection{Error Detection Prompt}
\label{app:error_detect_prompt}

We evaluate reasoning error detection using two benchmarks: ProcessBench and BIG-Bench-Mistake, each with its own prompting format.

For ProcessBench, we follow the original critique-based prompting scheme exactly as implemented in the official repository.\footnote{\url{https://github.com/QwenLM/ProcessBench}} The model is provided with a math problem and a paragraph-structured solution trace, and is instructed to identify the earliest paragraph containing an error, or return $-1$ if no error is present.

For BIG-Bench-Mistake, we follow the original task-specific templates provided by the authors.\footnote{\url{https://github.com/WHGTyen/BIG-Bench-Mistake}} We observe that strong reasoning models often solve the problem directly rather than performing error detection. To maintain the original task structure while enforcing focused evaluation, we prepend a minimal instruction requiring the model to identify incorrect steps only.

\medskip

\begin{promptbox}{ProcessBench Error Detection Prompt}
The following is a math problem and a solution (split into paragraphs, enclosed with tags and indexed from 0):

\medskip
\textbf{[Math Problem]}

\textcolor{cyan}{\{problem\}}

\medskip
\textbf{[Solution]}

\textcolor{cyan}{\{tagged\_response\}}

\medskip
Your task is to review and critique the solution paragraph by paragraph. Once you identify an error in a paragraph, return the index of the paragraph where the earliest error occurs. Otherwise, return the index of $-1$.

\medskip
Please put your final answer in \texttt{\textbackslash boxed\{\}}.
\end{promptbox}

\medskip

\begin{promptbox}{BIG-Bench-Mistake Task Template}

Given a problem and its solution steps below, identify whether there is any incorrect reasoning step. Do NOT solve the problem.

\medskip
\textbf{Task:}

\textcolor{cyan}{\{task\_description\}}

\medskip
\textbf{Problem:}

\textcolor{cyan}{\{input\}}

\medskip
\textbf{Reasoning:}

\textcolor{cyan}{\{steps\}}

\medskip
Return \texttt{\textbackslash boxed\{k\}} where $k$ is the step number of the incorrect step, or \texttt{\textbackslash boxed\{-1\}} if all steps are correct.

\end{promptbox}

\subsection{Test-time Scaling Prompt}
\label{app:tts_prompt}

We perform test-time scaling by injecting a corrupted chain-of-thought (CoT) into the model's intermediate thinking process and allowing the model to continue generating. This setup simulates partial erroneous reasoning and evaluates whether extended inference can recover from the injected mistake.

Concretely, each generation follows the structure:

\begin{promptbox}{Test-time Scaling Generation Format}

\textcolor{cyan}{\{question\}}

\medskip
$\langle\texttt{think}\rangle$

\medskip
\textcolor{red!100}{\{injected\_erroneous\_CoT\}}

\medskip
\textcolor{cyan}{\{continued\_reasoning\}}

\medskip
$\langle$\texttt{/think}$\rangle$

\medskip
\textcolor{cyan}{\{final\_answer\}}

\end{promptbox}

After the model completes its reasoning, we optionally append the token ``Wait'' multiple times to extend inference depth. This simple control mechanism encourages backtracking and revision of earlier steps, following prior work (e.g., \cite{muennighoff-etal-2025-s1, zeng-etal-2025-revisiting, ghosal2025does}).

\newpage
\section{Analysis: Error-Injected Intervention Variants}
\label{app:error_injected_variants}
In Section~\ref{sec:intervention_framework} and Figure~\ref{fig:intervention_framework}, we report the phenomenon in which LRMs often produce a correct final answer despite a corrupted CoT caused by injected errors ($\times$ Think, $\checkmark$ Answer). 
To probe how robust this phenomenon is, we design several \emph{error-injected variants} of the base intervention.
The injected error is identical across all variants; only the surrounding context changes:

\begin{itemize}
    \item \textbf{(a) Wait-once} (Figure~\ref{fig:variant_wait_once}). A single \emph{``Wait''} token is appended after the injected error, mimicking the test-time scaling cue of \citet{muennighoff-etal-2025-s1}. 
    This variant tests whether forcing the model to backtrack enables it to self-correct its reasoning.
    \item \textbf{(b) Misprint} (Figure~\ref{fig:variant_misprint}). We close \texttt{\textlangle/think\textrangle} immediately after the injected error, with no continued reasoning.
    This variant tests whether the model can recover from the injected error alone, before it corrupts the entire reasoning trace.
    \item \textbf{(c) Question-masked} (Figure~\ref{fig:variant_qmask}). The prompt is masked, so the model generates the final answer using only the corrupted CoT as context. 
    This setting tests if the model actually recovers from the error or simply ignores the CoT.
    \item \textbf{(d) Answer-prefix} (Figure~\ref{fig:variant_ansprefix}). The final-answer segment is forced to begin with \emph{``Based on my thinking above,''}.
    This instruction-prompting variant explicitly encourages the model to revisit and rely on the reasoning trace.
\end{itemize}

\subsection{Qualitative Examples}

We illustrate all variants using the same GSM8K example from Figure~\ref{fig:intervention_framework}.
All variants share the same question below, and use R1-32B with the arithmetic mistake $3+4=6$ injected after \texttt{\textlangle think\textrangle}:

\begin{tcolorbox}[colback=gray!8!white, colframe=gray!50!black, boxrule=0.3pt, arc=1mm, left=3mm, right=3mm, top=2mm, bottom=2mm]
\textit{Janet's ducks lay 16 eggs per day. She eats three for breakfast every morning and bakes muffins for her friends every day with four. She sells the remainder at the farmers' market daily for \$2 per fresh duck egg. How much in dollars does she make every day?} \hfill \textbf{\small (ground truth: \$18)}
\end{tcolorbox}

For brevity, each figure shows the shared question as a placeholder; variant~(c) instead masks the question entirely. Variant-specific cues are highlighted in \textbf{\textcolor{blue}{bold blue}}.

\begin{figure}[ht!]
\begin{promptbox}{(a) Wait-once -- R1-32B}

\textbf{Prompt:} \emph{[Janet's-ducks question, see above].}

\begin{tcolorbox}[colframe=red!60, colback=white, boxsep=1pt, title={}]
\texttt{\textlangle think\textrangle}\\[2pt]
First, I want to find how many eggs Janet uses before selling any. She eats 3 each morning and uses 4 more for muffins. \textcolor{txt_error}{So she uses $3+4=6$ eggs per day.} Next, I'll subtract from the 16 collected. \textcolor{txt_error}{So $16-6=10$ eggs are left. At \$2 each, $10\times\$2=\$20$. Janet makes \$20 every day.}

\textbf{\textcolor{blue}{Wait}}, that doesn't seem right. Let me double-check: she lays 16, uses 6, sells 10. \textcolor{txt_error}{At \$2 each, $10\times\$2=\$20$. So my final answer is \$20.}\\[2pt]
\texttt{\textlangle/think\textrangle}
\end{tcolorbox}

\begin{tcolorbox}[colframe=green!50!black, colback=white, boxsep=1pt, title={}]
\textbf{Final Answer:} Janet's ducks lay 16 eggs per day. She uses \textcolor{txt_correct}{$3$} for breakfast and \textcolor{txt_correct}{$4$} for muffins, totalling \textcolor{txt_correct}{$3+4=7$} eggs used. The remaining \textcolor{txt_correct}{$16-7=9$} eggs are sold at \$2 each, so she earns \textcolor{txt_correct}{$9\times\$2=\$18$} per day. \textcolor{txt_correct}{$\boxed{18}$}
\end{tcolorbox}

\end{promptbox}
\caption{\textbf{Wait-once.} The model fails to recover in the thinking trace, but still corrects the final answer. }
\label{fig:variant_wait_once}
\end{figure}

\begin{figure}[ht!]
\begin{promptbox}{(b) Misprint -- R1-32B}

\textbf{Prompt:} \emph{[Janet's-ducks question, see above].}

\begin{tcolorbox}[colframe=red!60, colback=white, boxsep=1pt, title={}]
\texttt{\textlangle think\textrangle}\\[2pt]
First, I want to find how many eggs Janet uses. She eats 3 each morning and uses 4 more for muffins. \textcolor{txt_error}{So she uses $3+4=6$ eggs per day.}\\[2pt]
\textbf{\textcolor{blue}{\texttt{\textlangle/think\textrangle}}}\quad \emph{\small(closed immediately, no continued reasoning)}
\end{tcolorbox}

\begin{tcolorbox}[colframe=green!50!black, colback=white, boxsep=1pt, title={}]
\textbf{Final Answer:} Janet's ducks lay 16 eggs per day. She uses \textcolor{txt_correct}{$3$} for breakfast and \textcolor{txt_correct}{$4$} for muffins, totalling \textcolor{txt_correct}{$3+4=7$} eggs used. The remaining \textcolor{txt_correct}{$16-7=9$} eggs are sold at \$2 each, so she earns \textcolor{txt_correct}{$9\times\$2=\$18$} per day. \textcolor{txt_correct}{$\boxed{18}$}
\end{tcolorbox}

\end{promptbox}
\caption{\textbf{Misprint.} Thinking stops immediately after the injected error and proceeds directly to the final answer.}
\label{fig:variant_misprint}
\end{figure}

\begin{figure}[ht!]
\begin{promptbox}{(c) Question-masked -- R1-32B}

\textbf{Prompt:} \textbf{\textcolor{blue}{[question masked]}}

\begin{tcolorbox}[colframe=red!60, colback=white, boxsep=1pt, title={}]
\texttt{\textlangle think\textrangle}\\[2pt]
First, I want to find how many eggs Janet uses. She eats 3 each morning and uses 4 more for muffins. \textcolor{txt_error}{So she uses $3+4=6$ eggs per day. Next, $16-6=10$ eggs are left. $10\times\$2=\$20$. Therefore, Janet makes \$20 every day.}\\[2pt]
\texttt{\textlangle/think\textrangle}
\end{tcolorbox}

\begin{tcolorbox}[colframe=green!50!black, colback=white, boxsep=1pt, title={}]
\textbf{Final Answer:} 16 eggs are collected each day. Janet uses \textcolor{txt_correct}{$3$} for breakfast and \textcolor{txt_correct}{$4$} for muffins, totalling \textcolor{txt_correct}{$3+4=7$} eggs used. The remaining \textcolor{txt_correct}{$16-7=9$} eggs are sold at \$2 each, so she earns \textcolor{txt_correct}{$9\times\$2=\$18$} per day. \textcolor{txt_correct}{$\boxed{18}$}
\end{tcolorbox}

\end{promptbox}
\caption{\textbf{Question-masked.} The question is masked, leaving only the corrupted reasoning trace as context.}
\label{fig:variant_qmask}
\end{figure}

\begin{figure}[ht!]
\begin{promptbox}{(d) Answer-prefix -- R1-32B}

\textbf{Prompt:} \emph{[Janet's-ducks question, see above].}

\begin{tcolorbox}[colframe=red!60, colback=white, boxsep=1pt, title={}]
\texttt{\textlangle think\textrangle}\\[2pt]
\textcolor{txt_error}{So she uses $3+4=6$ eggs per day. $16-6=10$ eggs left. $10\times\$2=\$20$. Therefore, Janet makes \$20.}\\[2pt]
\texttt{\textlangle/think\textrangle}
\end{tcolorbox}

\begin{tcolorbox}[colframe=green!50!black, colback=white, boxsep=1pt, title={}]
\textbf{Final Answer:} \textbf{\textcolor{blue}{Based on my thinking above,}} Janet's ducks lay 16 eggs per day. She uses \textcolor{txt_correct}{$3$} for breakfast and \textcolor{txt_correct}{$4$} for muffins, totalling \textcolor{txt_correct}{$3+4=7$} eggs used. The remaining \textcolor{txt_correct}{$16-7=9$} eggs are sold at \$2 each, so she earns \textcolor{txt_correct}{$9\times\$2=\$18$} per day. \textcolor{txt_correct}{$\boxed{18}$}
\end{tcolorbox}

\end{promptbox}
\caption{\textbf{Answer-prefix.} The model is instructed to revisit the thinking trace in its final response.}
\label{fig:variant_ansprefix}
\end{figure}

\subsection{Quantitative Results}

Table~\ref{tab:error_injected_variants_app} compares each variant against the \emph{intervened-local} baseline from Figure~\ref{fig:main_observation}, and the recovery phenomenon persists in every setting. Under \emph{Wait-once}, the rate falls on GSM8K because the explicit cue lets the model fix the error in plain text, but rises slightly on MATH-500, in line with overthinking on harder problems~\citep{chen2025do}. \emph{Misprint} attains the highest recovery rate, suggesting that even without error propagation through the rest of the CoT, the model self-corrects even more strongly. Surprisingly, under \emph{Question-masked}, the model still recovers at a rate close to the baseline despite having only the corrupted CoT as context. \emph{Answer-prefix} also lowers the rate slightly, but the phenomenon clearly persists for the larger models (R1-14B and R1-32B).

\begin{table*}[ht!]
\centering
\caption{Recovery rate ($\times$ Think, $\checkmark$ Final, in \%) across error-injected variants.}
\label{tab:error_injected_variants_app}
\begin{tblr}{
    width = \textwidth,
    colspec = {Q[l,m,4.5em] Q[l,m] *{5}{X[c,m]}},
    row{1} = {font=\bfseries, ht=2.4em},
    colsep = 4pt,
    rowsep = 1.2pt,
    rows = {ht = 1.5em},
}
\toprule
Dataset & Model & {Intervened\\-local} & {Wait\\-once} & Misprint & {Question\\-masked} & {Answer\\-prefix} \\
\midrule
\SetCell[r=4]{l} GSM8K
& Qwen3-4B & 20.12 & \phantom{0}3.03 & 80.52 & 12.13 & 12.21 \\
& R1-8B    & 55.12 & 21.76 & 67.32 & 41.39 & 31.61 \\
& R1-14B   & 77.43 & 31.46 & 77.63 & 62.02 & 55.27 \\
& R1-32B   & 74.00 & 30.25 & 80.14 & 66.79 & 62.62 \\
\midrule
\SetCell[r=4]{l} MATH-500
& Qwen3-4B & \phantom{0}5.80 & 11.60 & \phantom{0}8.40 & \phantom{0}5.80 & \phantom{0}5.40 \\
& R1-8B    & 41.00 & 48.80 & 41.60 & 33.00 & 35.80 \\
& R1-14B   & 58.20 & 59.60 & 58.00 & 43.80 & 32.60 \\
& R1-32B   & 53.80 & 55.80 & 53.80 & 46.20 & 32.20 \\
\bottomrule
\end{tblr}
\end{table*}

\subsection{Does the Model Simply Ignore the Wrong CoT?}
\label{app:ignore_cot}

An alternative explanation for the $\times$\,Think\,/\,$\checkmark$\,Final phenomenon is that the model simply discards the corrupted trace and re-solves the problem when generating the final answer:

\begin{tcolorbox}[colback=gray!10!white, colframe=gray!60!black, boxrule=0.4pt, arc=1mm, left=3mm, right=3mm, top=2mm, bottom=2mm]
\textit{Is the phenomenon simply caused by the model ignoring the corrupted thinking and solving the problem from scratch in the final answer?}
\end{tcolorbox}

The \emph{Question-masked} and \emph{Answer-prefix} variants are designed to test this hypothesis by forcing greater reliance on the corrupted trace.
In \emph{Question-masked}, the question is removed, leaving the corrupted CoT as the only available context. 
In \emph{Answer-prefix}, the final answer is explicitly prompted to refer back to the reasoning trace. 
In both cases, the $\times$\,Think\,/\,$\checkmark$\,Answer phenomenon persists, especially for larger models. 

\begin{takeaway}
The $\times$\,Think\,/\,$\checkmark$\,Answer phenomenon is not merely an artifact of ignoring the CoT; it persists even when the model is forced or instructed to condition on the corrupted trace.
\end{takeaway}

\FloatBarrier

\newpage
\section{Does Memorization Explain the Recovery Phenomenon?}
\label{app:memorization}
A key concern with the observation in Section~\ref{sec:hidden_recovery} is that the distilled model may have been trained on public benchmarks such as GSM8K. 
If so, the model might have memorized the final answers to these examples and could recover them without relying on the information contained in the corrupted reasoning trace. 
To rule out this possibility, we construct synthetic examples that do not appear in training and rerun the protocol from Section~\ref{sec:intervention_framework} on them:

\begin{itemize}
    \item[\hspace{1em}\textit{Step 1.}] For each GSM8K-test problem, we prompt GPT-5 to modify the numbers, names, and surface context while preserving the original step-by-step solution structure, yielding a new question with a single, consistent ground-truth answer.
    \item[\hspace{1em}\textit{Step 2.}] We retain only rewrites whose problem statement and answer are logically consistent.
    \item[\hspace{1em}\textit{Step 3.}] We apply the \emph{intervened-local} corruption protocol from Section~\ref{sec:intervention_framework} to the synthetic dataset and measure the same $\times$\,Think\,/\,$\checkmark$\,Final recovery rate.
\end{itemize}

The exact prompt template used in Step 1 is given below; we instantiate it once per GSM8K problem and query GPT-5 with structured JSON output.

\begin{promptbox}{Synthetic GSM8K Rewrite Prompt}
Create a new math problem and its ground truth answer based on the original
GSM8K sample.

\medskip
\textbf{RULES:} \\
- Change numbers, names, or context so the problem is new and not publicly
available online. \\
- Ensure the new numbers remain logical and consistent. \\
- Ensure the step-by-step reasoning is easy to follow. \\
- The final answer should be short and directly the true number.

\medskip
\textbf{Original Problem:} \\
Question: \textcolor{cyan}{\{question\}} \\
Ground Truth Answer: \textcolor{cyan}{\{answer\}}

\medskip
Return the result as JSON using the configured schema.
\end{promptbox}

An example of a synthetic sample:

\begin{tcolorbox}[colback=gray!5!white, colframe=gray!60!black, boxrule=0.4pt, arc=1mm, left=3mm, right=3mm, top=2mm, bottom=2mm]
\begin{tblr}{
    width = \linewidth,
    colspec = {Q[l,t,5em] X[l,t]},
    colsep = 4pt,
    rowsep = 3pt,
}
\textbf{Original} & \emph{Janet's ducks lay 16 eggs per day. She eats three for breakfast every morning and bakes muffins for her friends every day with four. She sells the remainder at the farmers' market daily for \$2 per fresh duck egg. How much in dollars does she make every day at the farmers' market?} \quad\textbf{Answer:} 18 \\
\textbf{Synthetic} & \emph{Carlos collects 30 oranges from his orchard each day. He keeps 6 for his family to eat and uses 8 to make fresh juice. He sells the remaining oranges at a market for \$1 each. How much money does Carlos earn per day from selling the oranges?} \quad\textbf{Answer:} 16 \\
\end{tblr}
\end{tcolorbox}

\paragraph{Results.}
Table~\ref{tab:memorization_synthetic} reports the recovery rate on the synthetic dataset alongside the original GSM8K-test numbers. Recovery rates remain large on the synthetic side -- $48.72\%$ for Qwen3-4B and above $68\%$ for all three R1 variants -- and stay close to the original GSM8K rates for the larger R1 models. Since the question, context, and final answer have all been altered and cannot have appeared in training or distillation data, this rules out memorization of the original answer as the explanation for the phenomenon.

\begin{table}[ht!]
\centering
\caption{Recovery rate ($\times$ Think, $\checkmark$ Final, in \%) of the original and synthetic intervention on GSM8K-test.}
\label{tab:memorization_synthetic}
\begin{tblr}{
    width = 0.7\textwidth,
    colspec = {Q[l,m] X[c,m] X[c,m]},
    row{1} = {font=\bfseries, ht=2.4em},
    colsep = 4pt,
    rowsep = 1.2pt,
    rows = {ht = 1.5em},
}
\toprule
Model & Original GSM8K-test & Synthetic GSM8K-test \\
\midrule
Qwen3-4B & 20.12 & 48.72 \\
R1-8B    & 55.12 & 68.54 \\
R1-14B   & 77.43 & 71.23 \\
R1-32B   & 74.00 & 71.58 \\
\bottomrule
\end{tblr}
\end{table}

\begin{takeaway}
The recovery in the final answer is not a memorized pattern from training data: the same $\times$\,Think\,/\,$\checkmark$\,Final phenomenon persists on new, unseen synthetic problems at comparable rates across all four models.
\end{takeaway}

\clearpage
\FloatBarrier
\newpage
\section{Additional Results of Logit Lens Interpretation} \label{app:logit_lens}

To validate the robustness of the identified steering vectors $v_\ell$, we analyze the trajectory of the top projected tokens with Logit Lens across five consecutive layers. Tables \ref{tab:logit_lens_qwen_appendix} and \ref{tab:logit_lens_r1-32b_split} present the evolution of the top 10 tokens and their associated logit scores for Qwen3-4B (Layers 29--33) and R1-32B (Layers 55--58), respectively. Comparing the two models reveals some qualitative differences in their self-critique mechanisms. While R1-32B (Table \ref{tab:logit_lens_r1-32b_split}) tends toward the interruptive tokens such as \textit{`Wait'} or `\cn{但实际上}' (but actually), Qwen3-4B (Table \ref{tab:logit_lens_qwen_appendix}) exhibits a preference for direct negation and correction tokens, such as \textit{`wrongly'}, \textit{`Nope'}, and \textit{`Actually'} throughout Layers 29--33. We hypothesize that this distinction may suggest that the extracted vector in Qwen3-4B encodes a more immediate ``fact-checking'' signal, whereas the vector in R1-32B captures a specialized backtracking behavior. This may also explain the significant difference in proportions of `$\times$ Think $\checkmark$ Answer' and `$\checkmark$ Think $\checkmark$ Answer' in Figure \ref{fig:main_observation} between the two models.

\begin{table}[ht!]
\caption{Evolution of top 10 predicted tokens and logit scores across five layers (Layer 29 to Layer 33) for the Qwen3-4B model.}
\label{tab:logit_lens_qwen_appendix}
\centering
\resizebox{\linewidth}{!}{%
\begin{tblr}{
    colspec = {Q[c,m] Q[l,m] Q[c,m] Q[l,m] Q[c,m] Q[l,m] Q[c,m] Q[l,m] Q[c,m] Q[l,m] Q[c,m]},
    colsep = 5pt,
    rowsep = 3pt,
    row{1-2} = {bg=gray!25},
    row{4,6,8,10,12} = {bg=gray!10},
}
    \toprule
    \SetCell[r=2]{c} \textbf{Rank} & \SetCell[c=2]{c} \textbf{Layer 29} & & \SetCell[c=2]{c} \textbf{Layer 30} & & \SetCell[c=2]{c} \textbf{Layer 31} & & \SetCell[c=2]{c} \textbf{Layer 32} & & \SetCell[c=2]{c} \textbf{Layer 33} & \\
    \cmidrule[lr]{2-3} \cmidrule[lr]{4-5} \cmidrule[lr]{6-7} \cmidrule[lr]{8-9} \cmidrule[lr]{10-11}
     & Token & Logit & Token & Logit & Token & Logit & Token & Logit & Token & Logit \\
    \midrule
    1  & wrongly   & 1.78 & {\cn{而且还} \scriptsize (And also)} & 2.28 & really & 2.49 & wrongly & 2.88 & !)\textbackslash n & 3.10 \\
    2  & Nope      & 1.72 & Nope & 2.21 & wrongly & 2.45 & !)\textbackslash n & 2.69 & wrongly & 2.91 \\
    3  & {\cn{而且还}  \scriptsize (And also)} & 1.63 & wrongly & 2.17 & {\cn{而且还} \scriptsize (And also)} & 2.42 & ---but & 2.65 & !)\textbackslash n\textbackslash n & 2.80 \\
    4  & really    & 1.61 & really & 2.11 & Nope & 2.38 & {\cn{而且还} \scriptsize (And also)} & 2.63 & )!\textbackslash n & 2.79 \\
    5  & \_that    & 1.60 & !)\textbackslash n & 2.02 & Actually & 2.22 & rethink & 2.62 & !!\textbackslash n\textbackslash n & 2.77 \\
    6  & Actually  & 1.60 & {\cn{只不过} \scriptsize (Just)} & 2.01 & !)\textbackslash n & 2.18 & !)\textbackslash n\textbackslash n & 2.62 & !!\textbackslash n & 2.75 \\
    7  & {\cn{但却}  \scriptsize (But yet)}   & 1.59 & Actually & 2.00 & {\cn{只不过} \scriptsize (Just)} & 2.18 & ?!\textbackslash n\textbackslash n & 2.55 & !!!\textbackslash n & 2.74 \\
    8  & ?)\textbackslash n\textbackslash n & 1.57 & !!!\textbackslash n & 1.99 & actually & 2.17 & !!!\textbackslash n & 2.52 & really & 2.74 \\
    9  & oque      & 1.57 & !)\textbackslash n\textbackslash n & 1.96 & !!!\textbackslash n & 2.15 & !!\textbackslash n\textbackslash n & 2.51 & ---but & 2.74 \\
    10 & actually  & 1.56 & choc & 1.94 & somebody & 2.15 & ---we & 2.49 & actually & 2.74 \\
    \bottomrule
\end{tblr}%
}
\end{table}

\begin{table}[ht!]
\caption{Evolution of top 10 predicted tokens and logit scores across four layers (Layer 55 to Layer 58) for the R1-32B model.}
\label{tab:logit_lens_r1-32b_split}
\centering
\resizebox{\linewidth}{!}{%
\begin{tblr}{
    colspec = {Q[c,m] Q[l,m] Q[c,m] Q[l,m] Q[c,m] Q[l,m] Q[c,m] Q[l,m] Q[c,m]},
    colsep = 5pt,
    rowsep = 3pt,
    row{1-2} = {bg=gray!25},
    row{4,6,8,10,12} = {bg=gray!10},
}
    \toprule
    \SetCell[r=2]{c} \textbf{Rank} & \SetCell[c=2]{c} \textbf{Layer 55} & & \SetCell[c=2]{c} \textbf{Layer 56} & & \SetCell[c=2]{c} \textbf{Layer 57} & & \SetCell[c=2]{c} \textbf{Layer 58} & \\
    \cmidrule[lr]{2-3} \cmidrule[lr]{4-5} \cmidrule[lr]{6-7} \cmidrule[lr]{8-9}
     & Token & Logit & Token & Logit & Token & Logit & Token & Logit \\
    \midrule
    1  & {\cn{但实际上} \scriptsize (But actually)} & 2.83 & {\cn{但实际上} \scriptsize (But actually)} & 2.66 & Wait & 2.68 & Wait & 2.95 \\
    2  & but & 2.54 & Wait & 2.60 & {\cn{但实际上} \scriptsize (But actually)} & 2.65 & Wait & 2.78 \\
    3  & Wait & 2.43 & Wait & 2.46 & {\cn{却} \scriptsize (But/Yet)} & 2.64 & however & 2.78 \\
    4  & {\cn{但由于} \scriptsize (But due to)} & 2.42 & however & 2.46 & however & 2.60 & ( & 2.74 \\
    5  & {\cn{但现在} \scriptsize (But now)} & 2.38 & {\cn{却} \scriptsize (But/Yet)} & 2.45 & but & 2.58 & {\cn{却} \scriptsize (But/Yet)} & 2.70 \\
    6  & {\cn{却被} \scriptsize (But was)} & 2.37 & but & 2.36 & {\cn{不过} \scriptsize (However)} & 2.57 & but & 2.68 \\
    7  & {\cn{却} \scriptsize (But/Yet)} & 2.30 & wait & 2.31 & Wait & 2.55 & wait & 2.65 \\
    8  & {\cn{但从} \scriptsize (But from)} & 2.28 & {\cn{但由于} \scriptsize (But due to)} & 2.30 & ( & 2.49 & {\cn{但实际上} \scriptsize (But actually)} & 2.62 \\
    9  & Wait & 2.18 & {\cn{却被} \scriptsize (But was)} & 2.28 & wait & 2.41 & {\cn{不过} \scriptsize (However)} & 2.58 \\
    10 & {\cn{不过} \scriptsize (However)} & 2.18 & However & 2.28 & However & 2.41 & However & 2.50 \\
    \bottomrule
\end{tblr}%
}
\end{table}

\newpage
\section{How to Select the Optimal Layer for Steering?}\label{app:layer_effect}
\begin{figure}[ht!]
    \centering
    \includegraphics[width=\textwidth]{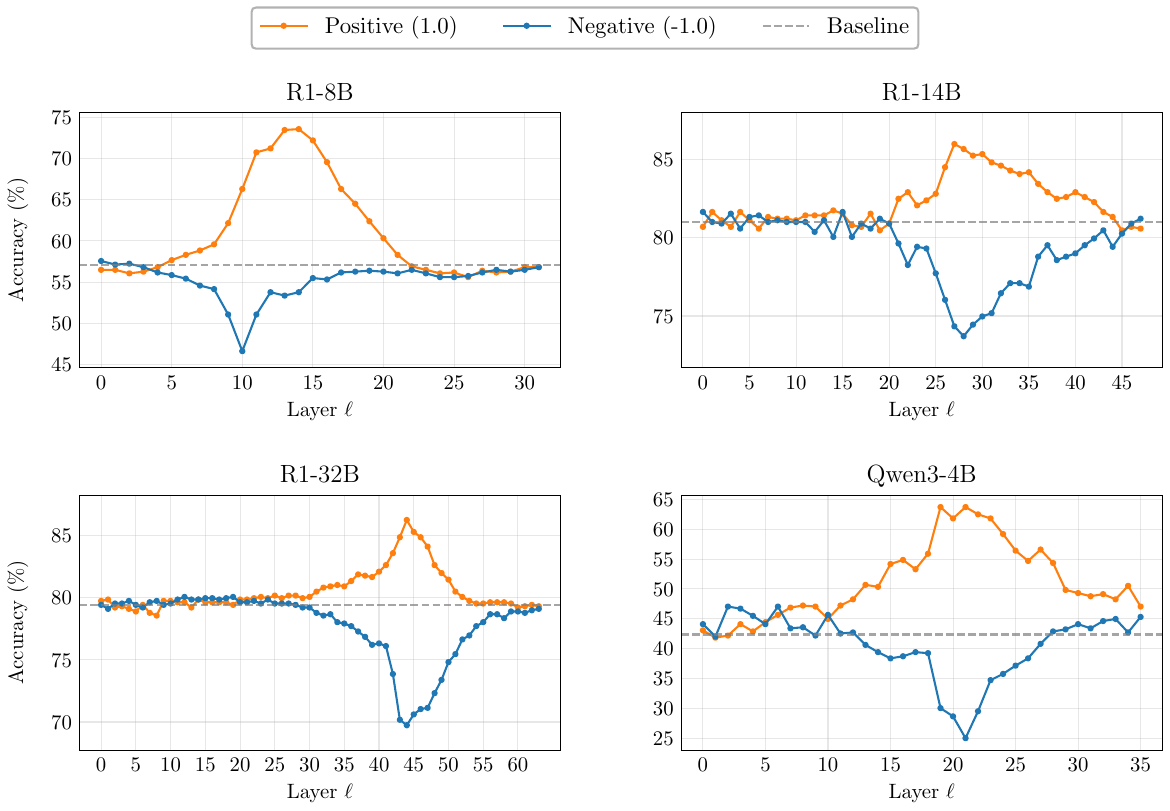}
    \caption{Layer-wise steering results. The separation between the performance of positive and negative steering is minimal for early and late layers. For middle layers, this performance gap is the most noticeable.}
    \label{fig:plot_layer_effect_app}
\end{figure}

Following previous works~\citep{zhao2025ahamomentsfakeidentifying, chen2025seal, venhoff2025understanding, sun2025thinkeditinterpretableweightediting}, we only apply the steering vector to the LRM for one selected layer. In this section, we describe the procedure for identifying the optimal layer that is the most representative of the self-critique mechanism. Consistent with~\citet{sun2025thinkeditinterpretableweightediting}, for each LRM, we apply the steering vector as defined in Eq. \ref{eq:steering_vector_addition} to each layer independently using scaling factors $\alpha \in \{-1.0,1.0\}$. All experiments for layer-wise steering are conducted on the GSM8K training split (Figure \ref{fig:plot_layer_effect_app}). We define the optimal layer as the one which yields the greatest accuracy differential between positive (1.0) steering and negative (-1.0) steering. At this layer, the effect of self-correction enhancement or suppression is the most pronounced, suggesting that this layer acts as the primary locus for the model's internal self-critique processes. As demonstrated in Table \ref{tab:optimal_layer_app} and Figure \ref{fig:plot_layer_effect_app}, the differential between positive and negative steering ($\Delta$) is the most prominent for middle layers, with $\Delta$ even exceeding $38\%$ for Layer 21 of Qwen3-4B.

\begin{table}[ht!]
    \centering
    \caption{Results of the optimal layer. Overall accuracy (\%) of the optimal layer evaluated on the training split of GSM8K for baseline ($\alpha=0.0$), positive steering ($\alpha=1.0$), negative steering ($\alpha=-1.0$), and the difference between positive and negative steering ($\Delta$).}
    \label{tab:optimal_layer_app}
    \begin{tblr}{
        colspec = {Q[l,6em] Q[c,5em] Q[c,5em] Q[c,5em] Q[c,5em]  Q[c,m]},
        row{1} = {font=\bfseries},
        colsep = 3pt,
        rowsep = 2pt,
        rows = {ht = 1.5em},
    }
        \toprule
        Model & \bm{$\alpha=0.0$} & \bm{$\alpha=-1.0$} & \bm{$\alpha=1.0$} & \bm{$\Delta$} & Layer \\
        \midrule
        Qwen3-4B & 42.36 & 25.00 & 63.72 & 38.72 & 21\\
        R1-8B & 57.03 & 53.36 & 73.48 & 20.13 & 13\\
        R1-14B & 80.99 & 73.71 & 85.64 & 11.93 & 28\\
        R1-32B & 79.42 & 69.74 & 86.25 & 16.51 & 44\\
        \bottomrule
    \end{tblr}%
\end{table}

\newpage
\section{Additional Results of Linear Separability of Internal Thoughts} \label{sec:activation_separability_app}

In Section \ref{sec:activation_separability}, we have provided the results of the best layer for the linear probing experiments. In this section, we outline the results (AUROC and ECE) for all layers across four LRMs: Qwen3-4B, R1-8B, R1-14B, and R1-32B.

\begin{figure}[ht!]
    \centering
    \includegraphics[width=\textwidth]{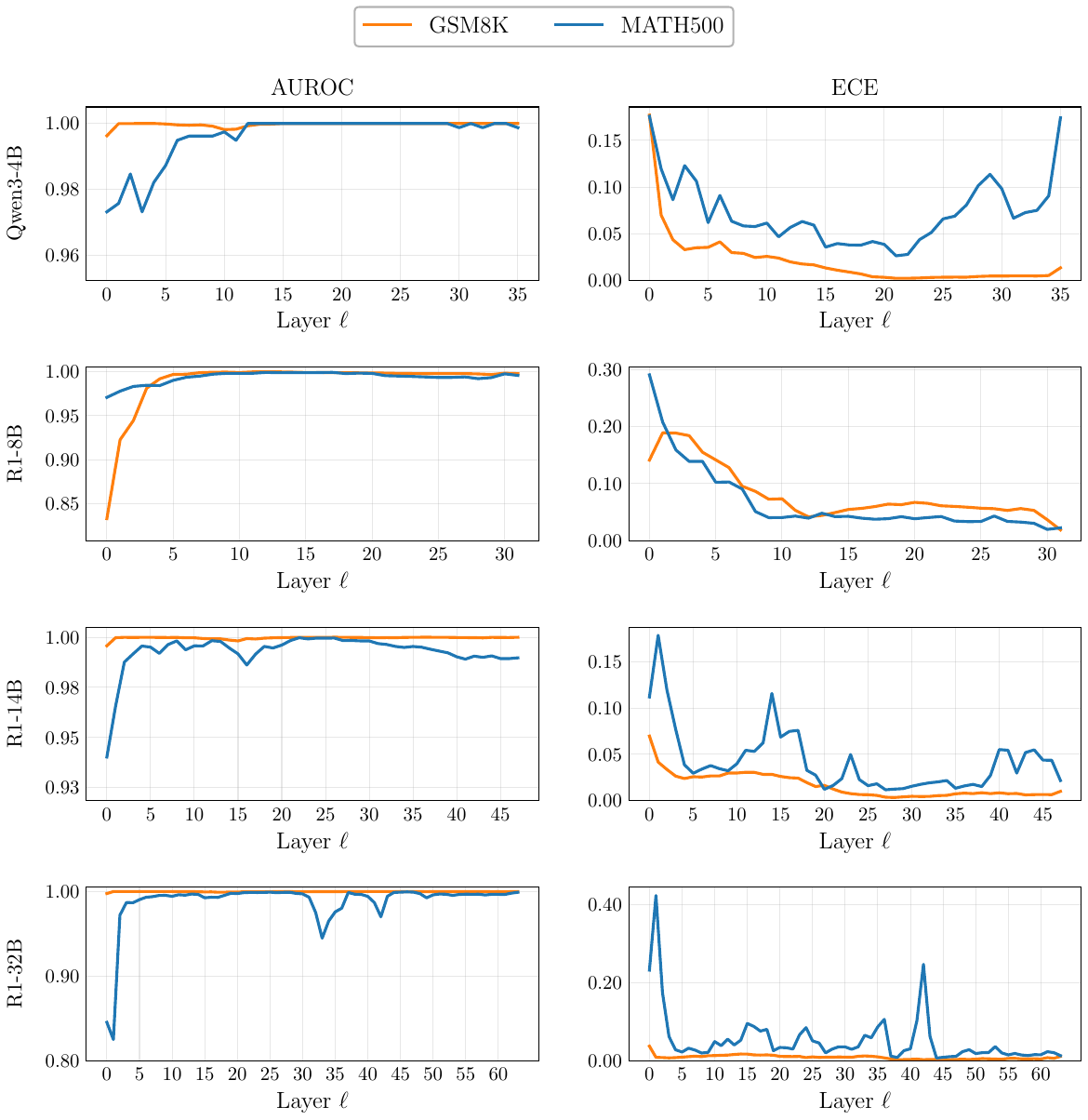}
    \caption{Layer-wise linear probing results. \textbf{Left to right:} AUROC and ECE for GSM8K (in-distribution) and MATH500 (out-of-distribution). For both settings, the linear prober can achieve near-perfect AUROC and low ECE at several layers.}
    \label{fig:linear_probing_results_app}
\end{figure}

As shown in the AUROC plots of Figure \ref{fig:linear_probing_results_app} (left column), linear separability saturates near 1.0 for the majority of the network depth across all models. The GSM8K dataset (in-distribution, orange) consistently exhibits high and stable separability almost immediately from the initial layers. In addition, we observe that lower layers tend to be poorly calibrated (high ECE), with calibration improving significantly in the middle layers. Comparing the two datasets, GSM8K demonstrates consistently lower Expected Calibration Error and smoother layer-wise progression compared to MATH500. The MATH500 dataset exhibits considerably higher variance and erratic spikes in ECE. However, for most layers, the performance of the linear prober for GSM8K and MATH500 remains close, suggesting this separability has a high level of generalization.

\clearpage
\section{Does the Probe Capture Critique, or Just Corrupted-Reasoning Style?}
\label{sec:probe_validity_app}

In Section~\ref{sec:activation_separability}, we trained a linear probe on baseline (unintervened, correct CoT) versus intervened-and-recovered (corrupted CoT) activations and observed near-perfect separability. A natural question follows:

\begin{tcolorbox}[colback=gray!10!white, colframe=gray!60!black, boxrule=0.4pt, arc=1mm, left=3mm, right=3mm, top=2mm, bottom=2mm]
\textit{Is this separability simply driven by an activation shift caused by the corrupted reasoning style introduced by error injection, rather than by the model's hidden critique behavior?}
\end{tcolorbox}

We address this concern with two analyses:
\begin{enumerate}
    \item \textbf{Project intervened-and-unrecovered samples onto the pretrained probe}
    (Section~\ref{sec:probe_projection_app}).
    \begin{itemize}
        \item \textit{What we do:} Similar to Eq.~\ref{eq:activation_extraction}, we extract the average activations of final-answer tokens from intervened samples that remain unrecovered ($\times$ Think, $\times$ Ans), and project them using the pretrained probe from Section~\ref{sec:activation_separability}.
        \item \textit{Expected result:} A high recovered-class rate would suggest that the probe mainly captures general CoT corruption style, which is shared across both groups. In contrast, a low rate would suggest that the probe reflects a hidden critique ability that distinguishes successful recovery from incorrect reasoning.
    \end{itemize}

    \item \textbf{Train a new probe on equally corrupted trajectories}
    (Section~\ref{sec:recov_unrecov_probe_app}).
    \begin{itemize}
        \item \textit{What we do:} We train a new probe from scratch using activations from recovered and unrecovered samples, both of which contain the same injected error style.
        \item \textit{Expected result:} If these activations are separable, this would suggest that the model contains an internal mechanism that distinguishes successful recovery from failure, rather than a difference caused merely by the corrupted CoT style.
    \end{itemize}
\end{enumerate}

\subsection{Project intervened-and-unrecovered samples onto the pretrained probe}
\label{sec:probe_projection_app}

\textbf{Setup.} 
We freeze the pretrained probe from Section~\ref{sec:activation_separability} and evaluate it on held-out GSM8K-test trajectories.
These trajectories are split into \emph{recovered} ($\times$ Think, $\checkmark$ Ans) and \emph{unrecovered} ($\times$ Think, $\times$ Ans), and their activations are projected onto the probe. 
We additionally evaluate two corruption variants unseen during probe training, \emph{Question-masked} (Figure~\ref{fig:variant_qmask}) and \emph{Misprint} (Figure~\ref{fig:variant_misprint}), as defined in Section~\ref{app:error_injected_variants}. 
For each population, we report the rate classified as \emph{recovered} (the positive class).

\begin{table}[ht!]
    \centering
    \caption{Rate (\%) of held-out activations classified as \emph{recovered} by the base probe of Section~\ref{sec:activation_separability}. A random classifier scores around $50\%$.}
    \label{tab:probe_projection_app}
    \begin{tblr}{
        width = \linewidth,
        colspec = {X[l,m,1.1] X[c,m,1.0] X[c,m,1.0] X[c,m,1.0] X[c,m,1.0] X[c,m,1.0] X[c,m,1.0]},
        colsep = 2pt,
        rowsep = 0.8pt,
        rows = {ht = 1.6em},
    }
        \toprule
        \SetCell[r=2]{c} \textbf{Model}
        & \SetCell[c=2]{c} \textbf{Original} &
        & \SetCell[c=2]{c} \textbf{Question masking} &
        & \SetCell[c=2]{c} \textbf{Misprinting} & \\
        \cmidrule[lr]{2-3} \cmidrule[lr]{4-5} \cmidrule[lr]{6-7}
        & Unrecovered & Recovered
        & Unrecovered & Recovered
        & Unrecovered & Recovered \\
        \midrule
        Qwen3-4B  & 64.22 & 99.53 & 67.58 & 99.38 & 65.27 & 100.00 \\
        R1-8B     & 67.38 & 99.78 & 66.00 & 99.45 & 69.96 &  99.61 \\
        R1-14B    & 68.49 & 98.65 & 45.97 & 99.76 & 68.64 & 100.00 \\
        R1-32B    & 67.82 & 99.75 & 33.71 & 99.88 & 67.82 &  99.80 \\
        \bottomrule
    \end{tblr}
\end{table}

\textbf{Results.} Recovered trajectories are classified as positive at very high rates, reaching at least $98.6\%$ across all models and unseen variants.
In contrast, unrecovered samples score only between $33.71\%$ and $69.96\%$, close to the $50\%$ binary baseline, despite sharing the same corrupted CoT style.
This indicates that the pretrained probe does not merely detect corrupted reasoning.
Rather, it decodes a positive signal only when the model successfully recovers, supporting the interpretation that the probe captures a hidden critique ability.

\subsection{Train a new probe on equally corrupted trajectories}
\label{sec:recov_unrecov_probe_app}

\textbf{Setup.}
We train a fresh logistic-regression probe to distinguish \emph{recovered} intervened trajectories ($\times$ Think, $\checkmark$ Ans) from \emph{unrecovered} intervened trajectories ($\times$ Think, $\times$ Ans). 
Both classes contain the same injected-error style, so the probe cannot rely on whether the CoT is corrupted. Instead, any separability must come from differences associated with the recovery outcome.
The probe is trained on GSM8K-train activations using class-balanced cross-entropy, since the two classes are not equally represented. 
For each model, we train at the steering layer used throughout the paper, selected by the procedure in Appendix~\ref{app:layer_effect}. 
We evaluate the resulting probe on GSM8K-test as the in-distribution (ID) setting, and on MATH500-test as the out-of-distribution (OOD) setting.

\begin{table}[ht!]
    \centering
    \caption{Linear separability of \emph{recovered} vs.\ \emph{unrecovered} intervened trajectories.}
    \label{tab:recov_unrecov_probe_app}
    \begin{tblr}{
        width = 0.78\linewidth,
        colspec = {X[l,m,1.2] X[c,m,0.7] X[c,m,1.0] X[c,m,1.0] X[c,m,1.0] X[c,m,1.0]},
        colsep = 2pt,
        rowsep = 0.8pt,
        rows = {ht = 1.6em},
    }
        \toprule
        \SetCell[r=2]{c} \textbf{Model}
        & \SetCell[r=2]{c} \textbf{Layer}
        & \SetCell[c=2]{c} \textbf{GSM8K} &
        & \SetCell[c=2]{c} \textbf{MATH500} & \\
        \cmidrule[lr]{3-4} \cmidrule[lr]{5-6}
        & & AUROC & ECE & AUROC & ECE \\
        \midrule
        Qwen3-4B & 21 & 0.9911 & 0.0262 & 0.9557 & 0.1154 \\
        R1-8B    & 13 & 0.9146 & 0.0676 & 0.8560 & 0.0580 \\
        R1-14B   & 28 & 0.8856 & 0.0732 & 0.8776 & 0.2532 \\
        R1-32B   & 44 & 0.9122 & 0.0771 & 0.8801 & 0.2524 \\
        \bottomrule
    \end{tblr}
\end{table}

\textbf{Results.} The recovered/unrecovered probe achieves strong separability across all models, with AUROC above $0.88$ on GSM8K-test and above $0.85$ on MATH500-test.
Calibration is very good on GSM8K-test, with ECE below $0.08$ for all models. On MATH500-test, ECE remains low for Qwen3-4B and R1-8B, but is higher for R1-14B and R1-32B; nevertheless, AUROC remains strong.
Since both classes share the same corrupted CoT style, this activation separability cannot be explained by injected errors alone. 
Instead, it suggests that when the model successfully recovers from wrong reasoning, its representations shift in a distinct way, which we later interpret as a decoded critique ability.

\begin{takeaway}
Even with equally corrupted CoTs, recovered and unrecovered runs remain separable in representation space. Thus, the linear probe from Section~\ref{sec:activation_separability} captures hidden recovery behavior, not merely a corrupted reasoning style.
\end{takeaway}

\FloatBarrier

\clearpage
\section{Comparison with Token Embedding Baseline}
\label{app:embedding_baseline}

A simple comparison for the critic vector from Section~\ref{sec:error_detection} is to steer instead with the input embedding of an error-related token. We use the token \texttt{wrong} -- the most direct lexical proxy for an error signal -- and take its input embedding $e_{\mathrm{wrong}} = \mathrm{embed}(\mathrm{tok}(\text{``wrong''}))$ as a steering vector.

\textbf{Setup.}
We inject $e_{\mathrm{wrong}}$ at the same steering layer used for the critic vector -- layer $21$ for Qwen3-4B, $13$ for R1-8B, $28$ for R1-14B, and $44$ for R1-32B, selected by the procedure in Appendix~\ref{app:layer_effect} -- with scaling factor $\alpha \in \{-1, +1\}$, and follow the protocol of Section~\ref{sec:error_detection}.
We evaluate both vectors on ProcessBench~\citep{zheng-etal-2025-processbench} and BIG-Bench Mistake~\citep{tyen-etal-2024-llms}, reporting Accuracy and F1.

\textbf{Results.}
Tables~\ref{tab:embedding_processbench} and~\ref{tab:embedding_bigbench} report each steering vector as a delta relative to the unsteered baseline.
The critic vector behaves as a clean axis: at $\alpha{=}-1$ Accuracy and F1 drop on every model on both benchmarks, and at $\alpha{=}+1$ they rise on every model on both benchmarks.
The \texttt{wrong}-embedding shows no such trend: its deltas are small and inconsistent in sign across models and benchmarks, so flipping its scale does not flip behavior the way a steering direction should.
This indicates that the critic vector is non-trivial to find: it is not recovered by simply taking the input embedding of an error-related token.

\begin{table}[ht!]
\centering
\caption{Accuracy and F1 (\%) on ProcessBench. $\alpha{=}0$ is the unsteered baseline.}
\label{tab:embedding_processbench}
\begin{tblr}{
    width = \textwidth,
    colspec = {Q[l,m] Q[l,m] *{3}{X[c,m]} *{3}{X[c,m]}},
    colsep = 5pt,
    rowsep = 0.6pt,
    rows = {ht = 1.4em},
    cell{1}{1} = {r=2}{c},
    cell{1}{2} = {r=2}{c},
    cell{1}{3} = {c=3}{c},
    cell{1}{6} = {c=3}{c},
}
    \toprule
    \textbf{Model} & \textbf{Method}
    & \textbf{Accuracy (\%)} & & 
    & \textbf{F1 (\%)} & & \\
    \cmidrule[lr]{3-5} \cmidrule[lr]{6-8}
    & & $\alpha{=}{-}1$ & $\alpha{=}0$ & $\alpha{=}{+}1$
    & $\alpha{=}{-}1$ & $\alpha{=}0$ & $\alpha{=}{+}1$ \\
    \midrule
    \SetCell[r=2]{l} Qwen3-4B
        & Ours   & \deltan{-4.19} & 78.69 & \deltap{+2.48} & \deltan{-4.85} & 80.26 & \deltap{+2.37} \\
        & Embed. & \deltan{-0.34} & --    & \deltap{+0.39} & \deltan{-0.38} & --    & \deltap{+0.38} \\
    \midrule
    \SetCell[r=2]{l} R1-8B
        & Ours   & \deltan{-1.48} & 40.46 & \deltap{+2.53} & \deltan{-1.94} & 41.16 & \deltap{+2.08} \\
        & Embed. & \deltan{-0.79} & --    & \deltan{-2.18} & \deltan{-0.99} & --    & \deltan{-2.54} \\
    \midrule
    \SetCell[r=2]{l} R1-14B
        & Ours   & \deltan{-1.57} & 72.08 & \deltap{+1.75} & \deltan{-1.82} & 73.14 & \deltap{+2.05} \\
        & Embed. & \deltan{-0.86} & --    & \deltap{+0.72} & \deltan{-0.95} & --    & \deltap{+0.46} \\
    \midrule
    \SetCell[r=2]{l} R1-32B
        & Ours   & \deltan{-2.69} & 75.70 & \deltap{+2.19} & \deltan{-3.25} & 77.27 & \deltap{+2.32} \\
        & Embed. & \deltan{-0.49} & --    & \deltan{-2.10} & \deltan{-0.65} & --    & \deltan{-2.07} \\
    \bottomrule
\end{tblr}
\end{table}

\begin{table}[ht!]
\centering
\caption{Accuracy and F1 (\%) on BIG-Bench Mistake. $\alpha{=}0$ is the unsteered baseline.}
\label{tab:embedding_bigbench}
\begin{tblr}{
    width = \textwidth,
    colspec = {Q[l,m] Q[l,m] *{3}{X[c,m]} *{3}{X[c,m]}},
    colsep = 5pt,
    rowsep = 0.6pt,
    rows = {ht = 1.4em},
    cell{1}{1} = {r=2}{c},
    cell{1}{2} = {r=2}{c},
    cell{1}{3} = {c=3}{c},
    cell{1}{6} = {c=3}{c},
}
    \toprule
    \textbf{Model} & \textbf{Method}
    & \textbf{Accuracy (\%)} & &
    & \textbf{F1 (\%)} & & \\
    \cmidrule[lr]{3-5} \cmidrule[lr]{6-8}
    & & $\alpha{=}{-}1$ & $\alpha{=}0$ & $\alpha{=}{+}1$
    & $\alpha{=}{-}1$ & $\alpha{=}0$ & $\alpha{=}{+}1$ \\
    \midrule
    \SetCell[r=2]{l} Qwen3-4B
        & Ours   & \deltan{-3.75} & 54.27 & \deltap{+3.59} & \deltan{-5.44} & 55.11 & \deltap{+4.60} \\
        & Embed. & \deltap{+0.16} & --    & \deltap{+1.90} & \deltap{+0.08} & --    & \deltap{+1.57} \\
    \midrule
    \SetCell[r=2]{l} R1-8B
        & Ours   & \deltan{-5.13} & 25.59 & \deltap{+7.32} & \deltan{-6.31} & 25.82 & \deltap{+7.47} \\
        & Embed. & \deltan{-5.04} & --    & \deltan{-1.82} & \deltan{-7.45} & --    & \deltan{-2.64} \\
    \midrule
    \SetCell[r=2]{l} R1-14B
        & Ours   & \deltan{-1.71} & 53.41 & \deltap{+1.26} & \deltan{-3.03} & 50.74 & \deltap{+1.20} \\
        & Embed. & \deltap{+0.40} & --    & \deltan{-0.30} & \deltap{+0.02} & --    & \deltan{-1.23} \\
    \midrule
    \SetCell[r=2]{l} R1-32B
        & Ours   & \deltan{-2.80} & 54.94 & \deltap{+2.33} & \deltan{-2.90} & 53.28 & \deltap{+3.40} \\
        & Embed. & \deltap{+3.72} & --    & \deltap{+0.65} & \deltap{+1.60} & --    & \deltap{+1.36} \\
    \bottomrule
\end{tblr}
\end{table}

\begin{takeaway}
Error-detection behavior cannot be steered by a simple token-embedding baseline; finding it requires the critic vector identified in this work.
\end{takeaway}

\FloatBarrier

\clearpage
\section{OOD Generalization of the Critic Vector}
\label{app:ood_generalization}

Throughout the main paper, the critic vector derived from GSM8K already demonstrates the controllability of critique behavior. Here we further test whether the same effect generalizes to a different source of training data.

\textbf{Setup.}
For each model, we recompute the critic vector from the MATH-500 training split following the exact procedure of Section~\ref{sec:error_detection}, and inject it at the same per-model steering layer (Appendix~\ref{app:layer_effect}). 
We then sweep $\alpha \in \{-1.0, -0.8, \ldots, +1.0\}$ and measure Accuracy and F1 on the Omnimath and OlympiadBench splits of ProcessBench, averaged. 
Both splits come from problem sources that are distinct from GSM8K and MATH-500, making them a clean OOD test and sufficient evidence that the vectors generalize beyond their source data.

\textbf{Results.}
Tables~\ref{tab:ood_processbench_acc} and~\ref{tab:ood_processbench_f1} show a consistent positive correlation between $\alpha$ and error-detection ability for both sources: $\alpha{>}0$ improves Accuracy and F1 on every model, and $\alpha{<}0$ degrades them. The MATH-500-sourced vector exerts a notably stronger pull on the larger models, with much sharper degradation under negative steering (e.g., $-16.15$ Acc / $-21.75$ F1 on R1-14B and $-20.15$ Acc / $-26.85$ F1 on R1-32B at $\alpha{=}-1.0$) and larger gains under positive steering (e.g., $+5.20$ Acc on R1-32B at $\alpha{=}+0.4$, $+5.00$ Acc / $+4.98$ F1 on R1-14B at $\alpha{=}+0.8$). This is consistent with the observation that steering vectors generalize better when derived from harder data~\citep{tan2024analysing}, and points to a promising direction: scaling the source data used to extract critique behavior. A thorough study of this scaling is beyond the scope of this work.

\begin{table}[ht!]
\centering
\small
\caption{Accuracy (\%) on ProcessBench under critic vectors sourced from GSM8K vs.\ MATH-500.}
\label{tab:ood_processbench_acc}
\footnotesize
\begin{tblr}{
    width = \textwidth,
    colspec = {Q[l,m,4.5em] Q[l,m,4.2em] *{5}{X[c,m]} X[c,m] *{5}{X[c,m]}},
    colsep = 3pt,
    rowsep = 0.6pt,
    rows = {ht = 1.5em},
}
    \toprule
    \textbf{Model} & \textbf{Source}
    & $-1.0$ & $-0.8$ & $-0.6$ & $-0.4$ & $-0.2$ & $\alpha{=}0$ & $+0.2$ & $+0.4$ & $+0.6$ & $+0.8$ & $+1.0$ \\
    \midrule
    \SetCell[r=2]{l} Qwen3-4B
        & GSM8K   & \deltan{-5.05} & \deltan{-3.90} & \deltan{-2.75} & \deltan{-2.20} & \deltan{-0.50} & \SetCell[r=2]{c} 73.45 & \deltan{-0.50} & \deltap{+1.70} & \deltap{+1.75} & \deltap{+1.60} & \deltap{+1.80} \\
        & MATH500 & \deltan{-4.90} & \deltan{-4.00} & \deltan{-3.10} & \deltan{-1.70} & \deltan{-2.65} &                        & \deltap{+0.35} & \deltap{+1.70} & \deltap{+1.90} & \deltap{+0.95} & \deltap{+2.40} \\
    \midrule
    \SetCell[r=2]{l} R1-8B
        & GSM8K   & \deltan{-3.55} & \deltan{-3.45} & \deltan{-1.70} & \deltan{-1.90} & \deltan{-0.75} & \SetCell[r=2]{c} 45.15 & \deltap{+1.80} & \deltap{+0.30} & \deltap{+1.55} & \deltap{+3.10} & \deltap{+4.85} \\
        & MATH500 & \deltan{-7.95} & \deltan{-7.75} & \deltan{-4.75} & \deltan{-3.10} & \deltan{-0.65} &                        & \deltap{+1.65} & \deltap{+1.65} & \deltap{+3.20} & \deltap{+2.30} & \deltap{+1.55} \\
    \midrule
    \SetCell[r=2]{l} R1-14B
        & GSM8K   & \deltan{-1.60} & \deltan{-1.35} & \deltan{-0.20} & \deltan{-1.45} & \deltan{-0.65} & \SetCell[r=2]{c} 65.15 & \deltap{+1.40} & \deltap{+1.50} & \deltap{+1.05} & \deltap{+2.05} & \deltap{+2.45} \\
        & MATH500 & \deltan{-16.15}& \deltan{-12.35}& \deltan{-9.60} & \deltan{-4.70} & \deltan{-2.30} &                        & \deltap{+2.50} & \deltap{+2.60} & \deltap{+3.50} & \deltap{+5.00} & \deltap{+4.85} \\
    \midrule
    \SetCell[r=2]{l} R1-32B
        & GSM8K   & \deltan{-2.05} & \deltan{-0.30} & \deltan{-1.20} & \deltap{+0.30} & \deltap{+0.25} & \SetCell[r=2]{c} 67.65 & \deltap{+0.80} & \deltap{+1.80} & \deltap{+1.45} & \deltap{+0.60} & \deltap{+2.95} \\
        & MATH500 & \deltan{-20.15}& \deltan{-15.80}& \deltan{-10.95}& \deltan{-7.25} & \deltan{-2.65} &                        & \deltap{+2.50} & \deltap{+5.20} & \deltap{+5.05} & \deltap{+2.95} & \deltap{+1.65} \\
    \bottomrule
\end{tblr}
\end{table}

\begin{table}[ht!]
\centering
\small
\caption{F1 (\%) on ProcessBench under critic vectors sourced from GSM8K vs.\ MATH-500.}
\label{tab:ood_processbench_f1}
\footnotesize
\begin{tblr}{
    width = \textwidth,
    colspec = {Q[l,m,4.5em] Q[l,m,4.2em] *{5}{X[c,m]} X[c,m] *{5}{X[c,m]}},
    colsep = 3pt,
    rowsep = 0.6pt,
    rows = {ht = 1.5em},
}
    \toprule
    \textbf{Model} & \textbf{Source}
    & $-1.0$ & $-0.8$ & $-0.6$ & $-0.4$ & $-0.2$ & $\alpha{=}0$ & $+0.2$ & $+0.4$ & $+0.6$ & $+0.8$ & $+1.0$ \\
    \midrule
    \SetCell[r=2]{l} Qwen3-4B
        & GSM8K   & \deltan{-5.25} & \deltan{-4.00} & \deltan{-2.79} & \deltan{-2.22} & \deltan{-0.47} & \SetCell[r=2]{c} 77.02 & \deltan{-0.72} & \deltap{+1.26} & \deltap{+1.19} & \deltap{+1.00} & \deltap{+1.10} \\
        & MATH500 & \deltan{-5.09} & \deltan{-4.12} & \deltan{-3.18} & \deltan{-1.67} & \deltan{-2.72} &                        & \deltap{+0.25} & \deltap{+1.54} & \deltap{+1.42} & \deltap{+0.30} & \deltap{+1.74} \\
    \midrule
    \SetCell[r=2]{l} R1-8B
        & GSM8K   & \deltan{-3.00} & \deltan{-3.10} & \deltan{-1.44} & \deltan{-1.70} & \deltan{-1.00} & \SetCell[r=2]{c} 46.64 & \deltap{+1.70} & \deltan{-0.19} & \deltap{+1.00} & \deltap{+2.30} & \deltap{+4.60} \\
        & MATH500 & \deltan{-10.17}& \deltan{-8.38} & \deltan{-4.53} & \deltan{-2.58} & \deltan{-0.48} &                        & \deltap{+1.30} & \deltap{+1.30} & \deltap{+1.46} & \deltap{+0.88} & \deltan{-0.43} \\
    \midrule
    \SetCell[r=2]{l} R1-14B
        & GSM8K   & \deltan{-1.88} & \deltan{-1.61} & \deltan{-0.30} & \deltan{-1.53} & \deltan{-0.68} & \SetCell[r=2]{c} 68.38 & \deltap{+1.44} & \deltap{+1.60} & \deltap{+1.15} & \deltap{+2.06} & \deltap{+2.56} \\
        & MATH500 & \deltan{-21.75}& \deltan{-15.94}& \deltan{-11.89}& \deltan{-5.68} & \deltan{-2.56} &                        & \deltap{+2.61} & \deltap{+2.71} & \deltap{+3.49} & \deltap{+4.98} & \deltap{+4.74} \\
    \midrule
    \SetCell[r=2]{l} R1-32B
        & GSM8K   & \deltan{-2.30} & \deltan{-0.32} & \deltan{-1.23} & \deltap{+0.34} & \deltap{+0.28} & \SetCell[r=2]{c} 71.02 & \deltap{+0.92} & \deltap{+1.86} & \deltap{+1.55} & \deltap{+0.67} & \deltap{+3.01} \\
        & MATH500 & \deltan{-26.85}& \deltan{-20.33}& \deltan{-13.33}& \deltan{-8.54} & \deltan{-2.86} &                        & \deltap{+2.52} & \deltap{+5.01} & \deltap{+4.98} & \deltap{+2.64} & \deltap{+1.33} \\
    \bottomrule
\end{tblr}
\end{table}

\begin{takeaway}
A critic vector derived from MATH-500 reproduces the controllability of critique behavior, confirming the framework is not tied to GSM8K. The stronger improvement from harder source data samples to a promising direction for scaling.
\end{takeaway}

\FloatBarrier

\clearpage
\section{Qualitative Examples}
\label{sec:qual_ex_app}

\subsection{Full Examples of Original and Intervened Generation}
\label{sec:generation_example_app}
In this section, we provide a complete example as shown in Figure \ref{fig:intervention_framework} from the GSM8K dataset generated by R1-32B (Figure \ref{fig:generation_example_xxx}).

\begin{figure}[!htbp]
\begin{tcolorbox}[title=Response Comparison - R1-32B, sidebyside, sidebyside align=top, colback=white, colframe=gray!60!white, boxsep=1pt, fontupper=\small, fontlower=\small,]
    \textbf{Original Response}
    \textbf{Prompt:} Janet's ducks lay 16 eggs per day. She eats three for breakfast every morning and bakes muffins for her friends every day with four. She sells the remainder at the farmers' market daily for \$2 per fresh duck egg. How much in dollars does she make every day at the farmers' market?\\

    \texttt{\textlangle think\textrangle} \\
    First, I need to determine the total number of eggs Janet's ducks lay each day. According to the problem, the ducks lay 16 eggs per day.

    Next, I'll calculate how many eggs Janet uses each day. \textcolor{txt_correct}{She eats 3 eggs for breakfast and uses 4 eggs for baking muffins, totaling 7 eggs consumed daily.}

    To find out how many eggs are left to sell, I'll subtract the eggs used from the total eggs laid: \textcolor{txt_correct}{16 eggs minus 7 eggs equals 9 eggs available for sale.}

    Finally, I'll calculate Janet's daily earnings from selling the eggs. \textcolor{txt_correct}{Each egg is sold for \$2, so multiplying the number of eggs sold by the price per egg gives 9 eggs multiplied by \$2, which equals \$18 per day.}\\
    \texttt{\textlangle /think\textrangle} \\

    \textbf{Final Answer:} Let's break down the problem step by step to find out how much Janet makes every day at the farmers' market.
    \begin{enumerate}
        \item \textbf{Total Eggs Laid Per Day:}
        \textcolor{txt_correct}{$16$ eggs/day}

        \item \textbf{Eggs Consumed Per Day:}
        \begin{itemize}
            \item \textcolor{txt_correct}{For Breakfast: $3$ eggs/day}
            \item \textcolor{txt_correct}{For Baking Muffins: $4$ eggs/day}
        \end{itemize}
        \textbf{Total Eggs Consumed:}
        \textcolor{txt_correct}{$3 + 4 = 7$ eggs/day}

        \item \textbf{Eggs Available for Sale:}
        \textcolor{txt_correct}{$16 - 7 = 9$ eggs/day}

        \item \textbf{Earnings from Selling Eggs:}
        \textcolor{txt_correct}{$9 \text{ eggs} \times \$2/\text{egg} = \$18/\text{day}$}
    \end{enumerate}

    \textbf{Final Answer:}
    \textcolor{txt_correct}{$\boxed{18}$}

    \tcblower

    \textbf{Intervened Response}
    \textbf{Prompt:} Janet's ducks lay 16 eggs per day. She eats three for breakfast every morning and bakes muffins for her friends every day with four. She sells the remainder at the farmers' market daily for \$2 per fresh duck egg. How much in dollars does she make every day at the farmers' market?\\

    \texttt{\textlangle think\textrangle}
    \begin{tcolorbox}[colframe=red!60, colback=white, boxsep=1pt, title={}]
    First, I want to find how many eggs Janet uses before selling any. \\
    She eats 3 each morning and uses 4 more for muffins. \\
    \textcolor{txt_error}{So she uses $3 + 4 = 6$ eggs per day.}
    \end{tcolorbox}

    Next, I'll subtract the eggs she uses from the total she collects.
    She collects 16 eggs each day.
    \textcolor{txt_error}{So, $16 - 6 = 10$ eggs are left to sell.}\\

    Finally, I'll calculate her earnings by multiplying the number of eggs she sells by the price per egg.
    She sells 10 eggs at \$2 each.
    \textcolor{txt_error}{So, $10 \times \$2 = \$20$.
    Therefore, Janet makes \$20 every day at the farmers' market.}\\
    \texttt{\textlangle /think\textrangle} \\

    \textbf{Final Answer:}
    Let's break down the problem step by step to find out how much Janet makes every day at the farmers' market.
    \begin{enumerate}
        \item \textbf{Total Eggs Laid Per Day:}
        \textcolor{txt_correct}{Janet's ducks lay a total of: $16 \text{ eggs/day}$}

        \item \textbf{Eggs Used by Janet:}
        \begin{itemize}
            \item \textcolor{txt_correct}{For Breakfast: Janet eats 3 eggs every morning.}
            \item \textcolor{txt_correct}{For Muffins: She uses 4 eggs daily.}
        \end{itemize}
        \textbf{Total Eggs Used:} \textcolor{txt_correct}{$3 + 4 = 7 \text{ eggs/day}$}

        \item \textbf{Eggs Available for Sale:} \textcolor{txt_correct}{$16 - 7 = 9 \text{ eggs/day}$}

        \item \textbf{Earnings from Selling Eggs:} \textcolor{txt_correct}{$9 \text{ eggs} \times \$2/\text{egg} = \$18$}
    \end{enumerate}

    \textbf{Final Answer:} \textcolor{txt_correct}{$\boxed{18}$}

\end{tcolorbox}

    \caption{Complete example shown in Figure \ref{fig:intervention_framework} generated by DeepSeek-R1-Distill-Qwen-32B (R1-32B). As laid out in Section \ref{sec:intervention_framework}, we generate the arithmetic error with GPT-5 (highlighted by a box in the right column) and insert this error immediately after the \texttt{$\langle$think$\rangle$} token, which causes the model to make subsequent mistakes (highlighted in \textcolor{txt_error}{red}) in the thinking stage. However, for the final answers, the model is able to recover from those thinking mistakes and derive the correct final solution. In this example, correct thinking steps and solutions are highlighted in \textcolor{txt_correct}{green}.}
    \label{fig:generation_example_xxx}
\end{figure}

\newpage
\subsection{Effects of Positive and Negative Steering on LRMs}
In this section, we present a qualitative analysis of R1-32B model's reasoning behaviors under the influence of steering vectors derived from the critique mechanism. We examine three distinct configurations: the Baseline Model (Coefficient 0.0), Positive Steering (Coefficient +1.0), and Negative Steering (Coefficient -1.0). The examples, selected from GSM8K samples of ProcessBench and GSM8K-Error, illustrate how steering modulates the model's capacity for error detection, semantic interpretation, and arithmetic verification. For this section, ellipsis is used with ``...'' to truncate model responses for brevity and clarity.

\subsubsection{Error Detection}\label{sec:error_qualitative_app}
We first examine how steering affects error detection capability, drawing on examples from ProcessBench-GSM8K and GSM8K-Error. These cases reveal two complementary failure modes that positive and negative steering address in opposite directions: missing genuine errors and hallucinating errors in correct solutions.

As shown in Figures~\ref{fig:qual_error_detection_gsm8k_app} and~\ref{fig:qual_ex_gsm8k}, positive steering enhances the model's ability to detect subtle reasoning errors while avoiding false positives. In contrast, Figure~\ref{fig:qual_ex_error_detection_gsm8k_negative_app} demonstrates that negative steering suppresses critical verification, causing the model to accept erroneous premises without scrutiny.
\begin{figure}[!ht]
    \centering
    
    \begin{tcolorbox}[
        paperbox,
        colback=bg_problem, 
        colframe=border_problem, 
        title=Problem Statement (ProcessBench - GSM8K),
        rounded corners
    ]
    \textbf{Problem:} Gerald works at a daycare that pays him \$30 every day. He worked for an \textit{entire week} and spent a total of \$100. How much does he have left?
    
    \vspace{0.3mm}
    \textbf{Target:} \$110 \quad \textbf{Question:} Identify any errors in the solution step below.
    
    \vspace{0.0cm}
    \hdashrule[0.5ex]{1.01\linewidth}{0.5pt}{2pt}
    
    \textit{Proposed Solution Step:}\\
    First, calculate the weekly earnings. Gerald earns \$30 each day. \textcolor{txt_error}{Assuming a standard workweek of 5 days}:
    \[ \text{Weekly earnings} = 5 \times \$30 = \$150 \]
    \end{tcolorbox}
    
    \begin{tcolorbox}[
        paperbox,
        colback=bg_baseline, 
        colframe=border_baseline, 
        title=Baseline Model (Coefficient 0.0),
        before skip=3pt
    ]
    The solution starts by calculating his weekly earnings. It mentions a standard workweek of 5 days. So, they multiply 5 by \$30, which gives \$150. \textcolor{txt_error}{That seems correct because 5 days times \$30 is indeed \$150.}
    
    \vspace{1mm}
    
    ... [Checks subtraction of expenses] ...
    
    \vspace{1mm}
    
    \textit{Conclusion:} The calculation steps seem accurate, and there doesn't appear to be any errors in the reasoning or arithmetic.
    \vspace{-1ex}
    \tcblower
    \vspace{-1ex} 
    \textbf{Outcome: The model overlooks the ``entire week'' wording and accepts the 5-day assumption.}
    \end{tcolorbox}
    
    \begin{tcolorbox}[
        paperbox,
        colback=bg_positive, 
        colframe=border_positive, 
        title=Positive Steering (Coefficient +1.0),
    ]
    First, the solution starts by calculating his weekly earnings... 5 by \$30 gives \$150. \textcolor{txt_correct}{But wait, the problem says he worked for an \textit{entire week}. Typically, a week has 7 days.} 
    
    \vspace{1mm}
    
    If he worked 7 days, his earnings should be $7 \times \$30 = \$210$. The initial solution assumes a 5-day workweek, but the problem implies the full week.
    
    \vspace{1mm}
    
    \textit{Conclusion:} The error is in assuming 5 days instead of 7.
    \vspace{-1ex}
    \tcblower
    \vspace{-1ex} 
    \textbf{\textcolor{txt_correct}{Outcome: Successfully identifies the reasoning error.}}
    \end{tcolorbox}
    
    \vspace{2mm}
    \caption{Qualitative example of positive steering on a GSM8K question from ProcessBench with R1-32B model. The proposed solution makes an unjustified assumption about the work schedule (5 days vs 7 days). The Baseline model accepts this assumption as ``standard'', while the steered model critically analyzes the specific phrasing ``entire week'' and identifies the discrepancy.}
    \label{fig:qual_error_detection_gsm8k_app}
    \end{figure}

    \begin{figure}[!ht]
    \centering
    
    \begin{tcolorbox}[
        paperbox,
        colback=bg_problem, 
        colframe=border_problem, 
        title=Problem Statement (GSM8K),
        rounded corners 
    ]
    \textbf{Problem:} Tom bought a CD for \$4, and a headphone set. In total, he paid \$48. How many more CDs would Tom have been able to buy if he had decided not to buy the headphone set?
    
    \vspace{0.3mm}
    \textbf{Proposed Solution:}
    ... To find the cost of the headphone set, we subtract the cost of the CD from the total amount he paid: $\$48 - \$4 = \$44$. So, the headphone set costs \$44.\\
    If Tom didn't buy the headphone set, he would have \$44 more to spend on CDs. Since each CD costs \$4, we can divide \$44 by \$4 to find out how many more CDs he could buy: $\$44 \div \$4 = 11$.
    \end{tcolorbox}
    
    \begin{tcolorbox}[
        paperbox,
        colback=bg_baseline, 
        colframe=border_baseline, 
        title=Baseline Model (Coefficient 0.0),
        before skip=3pt
    ]
    Let's check paragraph 2. "If Tom didn't buy the headphone set, he would have \$44 more..."
    Wait, the total amount he paid was \$48. If he didn't buy the headphones, he would have had the full \$48 available to spend on CDs. The solution claims he would have \$44 more, but \textcolor{txt_error}{he should have the total \$48. So 48 / 4 = 12 CDs. Since he already bought one, it's 11 more? No, the calculation of available funds is wrong.}
    
    \vspace{1mm}
    
    \textit{Conclusion:} The solution is incorrect because the available amount should be \$48.
    \vspace{-1ex}
    \tcblower
    \vspace{-1ex} 
    \textbf{Outcome: The model hallucinates an error in a correct solution (False Positive).}
    \end{tcolorbox}
    
    \begin{tcolorbox}[
        paperbox,
        colback=bg_positive, 
        colframe=border_positive, 
        title=Positive Steering (Coefficient +1.0),
    ]
    Checking paragraph 2... The problem states he bought a CD (\$4) and a headphone set. The total paid is \$48. This implies the headphone set cost $\$48 - \$4 = \$44$.
    If he decides \textit{not} to buy the headphone set, he saves exactly the cost of the headphones: \textcolor{txt_correct}{\$44}.
    He can then use this \$44 to buy more CDs: $\$44 / \$4 = 11$. This matches the solution.\\
    
    \textit{Conclusion:} The logic in paragraph 2 is sound. He saves \$44, not \$48, because the CD purchase is fixed.
    \vspace{-1ex}
    \tcblower
    \vspace{-1ex} 
    \textbf{\textcolor{txt_correct}{Outcome: Correctly validates the solution (Avoids Hallucination).}}
    \end{tcolorbox}
    
    \vspace{2mm}
    \caption{Qualitative example on GSM8K task with Qwen3-4B model. The baseline model hallucinates a logic error, incorrectly arguing that the total budget (\$48) should be used for the counterfactual, ignoring that the CD was still purchased. The positive steering helps the model successfully verify the correct logic: money saved equals the cost of the unpurchased item (\$44).}
    \label{fig:qual_ex_gsm8k}
    \end{figure}
\begin{figure}[!ht]
    \centering
    
    \begin{tcolorbox}[
        paperbox,
        colback=bg_problem, 
        colframe=border_problem, 
        title=Problem Statement (ProcessBench - GSM8K),
        rounded corners
    ]
    \textbf{Problem:} A caterer prepared hot dogs for 36 guests. He prepared enough for half of the guests to have two hotdogs (implying the other half has one). However, 40 guests showed up, and everyone wanted a second hotdog.
    
    \vspace{0.3mm}
    \textbf{Question:} How many guests did not get a second hotdog?
    
    \vspace{0.0cm}
    \hdashrule[0.5ex]{1.01\linewidth}{0.5pt}{2pt}
    
    \textit{Proposed Solution Step 1:}\\
    The caterer planned for half of 36 guests (18) to eat two hotdogs. \textcolor{txt_error}{So, the total number of hotdogs prepared is $18 \times 2 = 36$ hotdogs.} (Note: Ignores the single hotdogs for the other 18 guests).
    \end{tcolorbox}
    
    \begin{tcolorbox}[
        paperbox,
        colback=bg_baseline,
        colframe=border_baseline, 
        title=Baseline Model (Coefficient 0.0),
        before skip=3pt,
    ]
    Let's figure out how many hotdogs he prepared. Half of 36 is 18, so 18 guests have two hotdogs ($18 \times 2 = 36$). \textbf{Wait, hold on a second.} If half are having two, the other half would have one each. So the total should be $36 + 18 = 54$ hotdogs. \textcolor{txt_correct}{But the solution says he prepared 36 hotdogs. That doesn't seem right.}
    
    \vspace{1mm}
    
    The solution only accounted for the 36... their initial calculation is wrong.
    \vspace{-1ex}
    \tcblower
    \vspace{-1ex}
    \textbf{Outcome: Correctly identifies the error (54 vs 36 prepared).}
    \end{tcolorbox}
    
    \begin{tcolorbox}[
        paperbox,
        colback=bg_negative, 
        colframe=border_negative, 
        title=Negative Steering (Coefficient -1.0),
    ]
    Half of 36 is 18, so 18 guests would have two hotdogs each. Multiplying 18 by 2 gives 36 hotdogs. \textcolor{txt_error}{So, the caterer prepared 36 hotdogs in total.}
    
    \vspace{2pt}
    ...If the caterer only prepared 36 hotdogs, and 40 guests wanted a second... 4 guests couldn't get a second. The solution seems accurate... Therefore, the solution is correct, and there's no error.
    \vspace{-1ex}
    \tcblower
    \vspace{-1ex}
    \textbf{\textcolor{txt_error}{Outcome: Fails to spot the error (Maintains the Error).}}
    \end{tcolorbox}
    
    \vspace{2mm}
    \caption{Qualitative example of negative steering on ProcessBench-GSM8K with R1-32B model. The prompt contains a logical error in the first step (calculating total food prepared). The baseline model successfully debugs the setup ($54$ hotdogs vs $36$), whereas negative steering suppresses this critical check and accepts the erroneous premise.}
    \label{fig:qual_ex_error_detection_gsm8k_negative_app}
\end{figure}

\clearpage
\subsubsection{Test-Time Scaling}\label{sec:tts_qualitative_app}
We next examine how steering interacts with the model's internal self-verification behavior with test-time scaling methods.

Figures~\ref{fig:qual_ex_tts_gsm8k_stack_1_app} and~\ref{fig:qual_ex_tts_gsm8k_negative_app} illustrate that positive steering enables successful error recovery during the ``Wait'' phase, whereas negative steering causes the model to rationalize and reinforce injected errors rather than correct them.
\begin{figure}[!ht]
    \centering
    
    \begin{tcolorbox}[
        paperbox,
        colback=bg_problem, 
        colframe=border_problem, 
        title=Problem Statement (GSM8K-Error),
        rounded corners 
    ]
    \textbf{Question:} 
    Bailey starts with a certain amount of money. Then she receives a weekly allowance of \$5 for 8 weeks. At the end of the 8 weeks, if she has a total of \$100, how much money did Bailey start with?
    
    \vspace{0.3mm}
    \textbf{Target:} 60 
    
    \vspace{0.0cm}
    \hdashrule[0.5ex]{1.01\linewidth}{0.5pt}{2pt}
    
    \textit{Injected Error:} 
    
    First, I need to find the total amount she received from allowance: $\$5$ each week for $8$ weeks.\\
    So I’ll multiply \textcolor{txt_error}{$5 × 8 = 30$} and subtract it from the final total ($\$100$).
    \end{tcolorbox}
    
    \begin{tcolorbox}[
        paperbox,
        colback=bg_baseline, 
        colframe=border_baseline, 
        title=Baseline Model (Coefficient 0.0),
        before skip=3pt
    ]
    
    
    Wait, that doesn’t make sense... But just to double-check, let me add \$70 and \$30 again... \textcolor{txt_error}{Yes, $70 + 30 = 100$. So, my answer is correct.}
    \vspace{-1ex}
    \tcblower
    \vspace{-1ex} 
    \textbf{Outcome: The model fails to spot the arithmetic error ($5 \times 8 \neq 30$).}
    \end{tcolorbox}
    
    \begin{tcolorbox}[
        paperbox,
        colback=bg_positive, 
        colframe=border_positive, 
        title=Positive Steering (Coefficient +1.0),
    ]
    
    Wait, but let me double-check. If she started with \$70 and received \$5 each week for 8 weeks... \textcolor{txt_correct}{Wait, $5 \times 8$ is actually 40, not 30.}
    
    So initial money should be $100 - 40 = 60$. Yes, that makes sense. If she started with \$60 and received \$5 for 8 weeks, that’s $60 + 40 = 100$.\\
    
    \textit{Conclusion:} The initial calculation of allowance was wrong.
    \vspace{-1ex}
    \tcblower
    \vspace{-1ex} 
    \textbf{\textcolor{txt_correct}{Outcome: Identifies \& Fixes Error ($30 \rightarrow 40$).}}
    \end{tcolorbox}
    
    \vspace{2mm}
    \caption{Qualitative example of positive steering on GSM8K-Error with R1-32B model. The baseline model uses a multiplication error ($5 \times 8 = 30$) and confirms it during self-verification ($70+30=100$). Positive steering initially makes the same error but successfully catches it during the ``Wait'' phase, correcting the allowance to $40$ and the final answer to $60$.}
    \label{fig:qual_ex_tts_gsm8k_stack_1_app}
    \end{figure}

\begin{figure}[!ht]
\centering

\begin{tcolorbox}[
    paperbox,
    colback=bg_problem, 
    colframe=border_problem, 
    title=Problem Statement (GSM8K-Error),
    rounded corners
]
\textbf{Question:} Juan and Carlos sell lemonade. Costs/gallon: \$3 lemons, \$2 sugar. Sell: 20 glasses at \$0.50. Profit \$25. How much spent on lemons?

\vspace{0.0cm}
\hdashrule[0.5ex]{1.01\linewidth}{0.5pt}{2pt}

\textit{Injected Error:}\\
The costs are \$3 for lemons and \$2 for sugar per gallon, so \textcolor{txt_error}{the total cost per gallon is \$6}. Revenue per gallon is \$10. Profit per gallon is \$4.
If total profit is \$25, gallons sold is 6.25.
Total spent on lemons is 6.25 * \$3 = \$18.75.
\end{tcolorbox}

\begin{tcolorbox}[
    paperbox,
    colback=bg_baseline,
    colframe=border_baseline, 
    title=Baseline Model (Coefficient 0.0),
    before skip=3pt,
]
Wait, but the problem asks for the amount spent on lemons... So, 6.25 gallons * \$3 = \$18.75.

\vspace{1mm}

\textbf{Wait, but let me double-check my steps to make sure I didn't make a mistake.}
Total cost per gallon: \$3 (lemons) + \$2 (sugar) = \$5? \textcolor{txt_correct}{Wait, no, \$3 + \$2 is \$5, so total cost per gallon is \$5, not \$6.}
...
Therefore, the amount spent on lemons is 5 gallons * \$3 per gallon = \$15.
\vspace{-1ex}
\tcblower
\vspace{-1ex}
\textbf{Outcome: The model catches the addition error (\$3+\$2=\$5) and corrects the answer.}
\end{tcolorbox}

\begin{tcolorbox}[
    paperbox,
    colback=bg_negative, 
    colframe=border_negative, 
    title=Negative Steering (Coefficient -1.0),
]
Wait, but the problem asks for the amount spent on lemons... So, using the number of gallons sold, which is 6.25, multiplied by the cost per gallon for lemons, which is \$3, gives the total amount spent on lemons.

So, \textcolor{txt_error}{6.25 gallons * \$3 per gallon = \$18.75.}

But wait, the problem might expect an exact answer... 25/4 * 3 = 75/4, which is \$18.75.
Yes, that makes sense.
\vspace{-1ex}
\tcblower
\vspace{-1ex}
\textbf{\textcolor{txt_error}{Outcome: Maintains the Error.}}
\end{tcolorbox}

\vspace{2mm}
\caption{Qualitative example of negative steering on ProcessBench-GSM8K with R1-32B model. The prompt context introduces an arithmetic error (\$3+\$2=\$6 instead of \$5). The baseline model initially adopts the error but triggers a self-correction after generating ``Wait,'' re-evaluating the costs. The negatively steered model accepts the erroneous premise and justifies the incorrect result.}
\label{fig:qual_ex_tts_gsm8k_negative_app}
\end{figure}

\end{document}